\documentclass[a4paper,fleqn]{cas-dc}

\usepackage{algorithm}
\usepackage{algorithmic}
\usepackage{graphicx}
\usepackage[algo2e]{algorithm2e} 
\usepackage[round]{natbib}
\usepackage{subcaption}
\usepackage{pifont}
\usepackage{bbm}
\usepackage{xcolor}
\usepackage{graphicx}
\usepackage{multirow}
\usepackage{bbm}
\usepackage{enumitem}
\usepackage{acronym}

\usepackage[normalem]{ulem}
\useunder{\uline}{\ul}{}
\AtEndPreamble{
    \usepackage[capitalize]{cleveref}
    \crefname{section}{Sec.}{Secs.}
    \Crefname{section}{Section}{Sections}
    \Crefname{table}{Table}{Tables}
    \crefname{table}{Tab.}{Tabs.}
}
\newcommand{\cy}[1]{\textcolor{black}{#1}}

\def\tsc#1{\csdef{#1}{\textsc{\lowercase{#1}}\xspace}}
\tsc{WGM}
\tsc{QE}

\usepackage[switch]{lineno}

\begin{document}
\let\WriteBookmarks\relax
\def\floatpagepagefraction{1}
\def\textpagefraction{.001}
\captionsetup[figure]{labelfont={bf},labelformat={default},labelsep=period,name={Fig.}}

\shorttitle{LandSegmenter}    

\shortauthors{Chenying Liu, et al.}  

\title [mode = title]{{LandSegmenter: Towards a Flexible Foundation Model for Land Use and Land Cover Mapping}}



%


\author[1,2]{Chenying Liu}[type=editor,
					  auid=2,
					  bioid=2,
					  prefix=,]
\ead{chenying.liu@tum.de}

\author[1]{Wei Huang}[type=editor,
			          auid=1,
			          bioid=1,
			          prefix=,]
\ead{w2wei.huang@tum.de}

\author[1,2]{Xiao Xiang Zhu}[type=editor,
					  auid=5,
					  bioid=5,
					  prefix=,]
\ead{xiaoxiang.zhu@tum.de}
\cormark[1]






\affiliation[1]{organization={Chair of Data Science in Earth Observation, Technical University of Munich},
				city={Munich},
				postcode={80333},
				country={Germany}}
				
\affiliation[2]{organization={Munich Center for Machine Learning (MCML)},
				city={Munich},
				postcode={80333},
				country={Germany}}








\cortext[1]{Corresponding author}


\newacro{AFM}{Attention-based Fusion Module}
\newacro{CLIP}{Contrastive Language-Image Pretraining}
\newacro{CV}{Computer Vision}
\newacro{EO}{Earth Observation}
\newacro{FMs}{Foundation Models}
\newacro{GEE}{Google Earth Engine}
\newacro{GT}{ground-truth}
\newacro{LAS}{LAnd Segment}
\newacro{LULC}{Land Use and Land Cover}
\newacro{MAE}{Masked Autoencoders}
\newacro{MIM}{Masked Image Modeling}
\newacro{OVSS}{Open-Vocabulary Semantic Segmentation}
\newacro{RS}{Remote Sensing}
\newacro{SAM}{Segment Anything Model}
\newacro{SSL}{Self-Supervised Learning}
\newacro{VFM}{Vision Foundation Model}
\newacro{VLMs}{Vision-Language Models}
\newacro{VQA}{Visual Question Answering}

\begin{abstract}
Land Use and Land Cover (LULC) mapping is a fundamental task in Earth Observation (EO). However, current LULC models are typically developed for a specific modality and a fixed class taxonomy, limiting their generability and broader applicability. Recent advances in foundation models (FMs) offer promising opportunities for building universal models. Yet, task-agnostic FMs often require fine-tuning for downstream applications, whereas task-specific FMs rely on massive amounts of labeled data for training, which is costly and impractical in the remote sensing (RS) domain.
To address these challenges, we propose LandSegmenter, an LULC FM framework that resolves three-stage challenges at the input, model, and output levels. From the input side, to alleviate the heavy demand on labeled data for FM training, we introduce LAnd Segment (LAS), a large-scale, multi-modal, multi-source dataset built primarily with globally sampled weak labels from existing LULC products. LAS provides a scalable, cost-effective alternative to manual annotation, enabling large-scale FM training across diverse LULC domains. For model architecture, LandSegmenter integrates an RS-specific adapter for cross-modal feature extraction and a text encoder for semantic awareness enhancement. 
At the output stage, we introduce a class-wise confidence-guided fusion strategy to mitigate semantic omissions and further improve LandSegmenter's zero-shot performance. We evaluate LandSegmenter on six precisely annotated LULC datasets spanning diverse modalities and class taxonomies. Extensive transfer learning and zero-shot experiments demonstrate that LandSegmenter achieves competitive or superior performance, {particularly in zero-shot settings when transferred to unseen datasets.} These results highlight the efficacy of our proposed framework and the utility of weak supervision for building task-specific FMs. {The code and dataset are publicly available at \url{https://github.com/zhu-xlab/LandSegmenter.git}.}
\end{abstract}




\maketitle

\section{Introduction}
\label{sec:introduction}

\ac{LULC} mapping is critical for many real \ac{EO} applications~\citep{chen_2025_superpixel, wang_review_2023, shi_efficient_2025}. 
Joint efforts from \ac{RS} and \ac{CV} communities have driven significant advances in LULC mapping using traditional handcrafted and modern deep-learning-based approaches~\citep{zhu_deep_2017,he_recent_2018,prudente_multisensor_2022}. However, these models are often tailored to specific geographic regions, data types, and tasks, constraining generalizability and scalability for large-scale deployment~\citep{chen_toward_2023}.

Recent developments in \ac{FMs} offer a promising avenue for addressing these limitations. Task-agnostic \ac{FMs} leverage \ac{SSL}~\citep{he_momentum_2020,he_masked_2022,caron_emerging_2021} to enhance feature representation from unlabeled data, enabling efficient adaptation to downstream tasks with minimal labeled data~\citep{wang_self-supervised_2022}. Conversely, task-specific FMs utilize extensive labeled data to achieve robust performance for specialized purposes, exemplified by the \ac{SAM} series~\citep{kirillov_segment_2023,ravi_sam_2024} for promptable segmentation. These advancements suggest the potential of \ac{FMs} to provide unified and scalable solutions. 

However, applying \ac{SAM} to \ac{LULC} mapping presents several challenges due to unique \ac{RS} data characteristics, including diverse modalities, varying spatial resolutions, and domain-specific features. \ac{SAM} models were trained on high-resolution RGB natural images and videos. They struggle with multispectral \ac{RS} data, which are often of medium to low spatial resolution, such as Sentinel (10m) and Landsat (30m). While \ac{SAM} performs well on instance segmentation of discrete, well-bounded objects (e.g., cars), it falls short when handling region-level classes (e.g., grass) that represent continuous or less distinctly bounded land surfaces~\citep{ji_segment_2024, Zhu_2025_CVPR} and are essential for accurate \ac{LULC} mapping. Its reliance on geometric prompts (e.g., points, boxes) also limits its effectiveness in dense mapping.

On the other hand, training task-specific \ac{FMs} demands extensive labeled data. Precise labels are costly and labor-intensive. For instance, SAM training relies on millions of images and over a billion annotated masks~\citep{kirillov_segment_2023}, a scale impractical for RS applications. Weakly supervised pretraining offers a solution by utilizing abundant albeit imperfect labels~\citep{mahajan_exploring_2018,ghadiyaram_large-scale_2019,jia_scaling_2021}. While some studies~\citep{maggiori_convolutional_2017, liu_cromss_2025} have explored pixel-wise weak labels for pretraining, they primarily focus on simpler or smaller-scale tasks. Its potential in complex, large-scale dense prediction tasks remains underexplored. 
{LULC mapping, with widely available noisy products easily paired with RS imagery, offers an ideal test case. Although these products contain label noise from automatic errors, ambiguities, and temporal inconsistencies, they can provide sufficient supervision for models to capture dominant spatial patterns while remaining fairly robust to noise.}

\begin{figure*}
    \centering
    \begin{tabular}{ccc}
    \includegraphics[width=.3\textwidth]{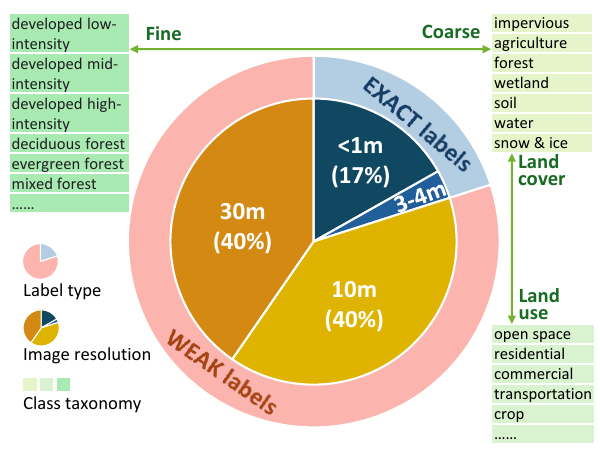} &
    \includegraphics[width=.3\textwidth]{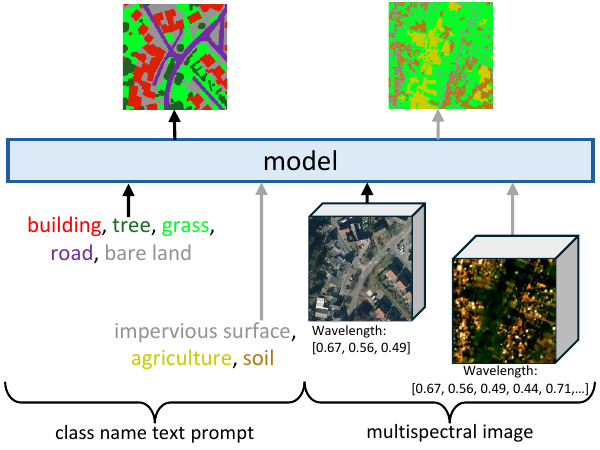} &
    \includegraphics[width=.3\textwidth]{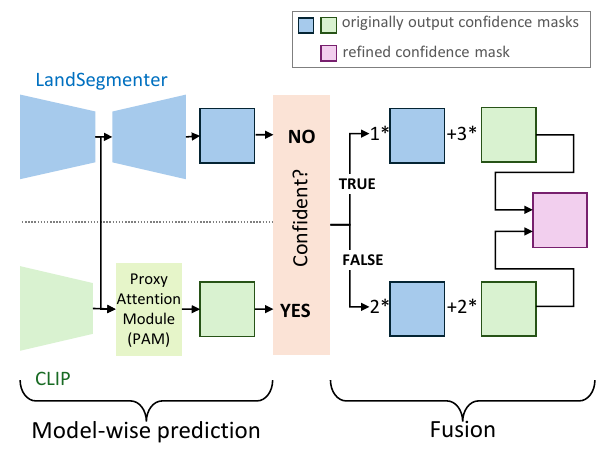} \\
    (a) {\textbf{Input}: LAnd Segment (LAS) dataset} & (b) \textbf{Model}: LandSegmenter & (c) \textbf{Inference}: Confidence-guided fusion \\
    \end{tabular}
    \caption{Overview of the proposed workflow for LULC FM construction, comprising three main stages. (a) LAS dataset curation: a globally sampled collection of RS imagery spanning diverse {modalities} and LULC categories, {\ul primarily weakly labeled at low cost}. (b) LandSegmenter model design: a task-adaptive architecture capable of processing varying multispectral inputs and producing LULC maps tailored to user-defined category sets. (c) Zero-shot inference enhancement: a confidence-guided fusion strategy to improve recognition of semantically omitted or underrepresented classes during inference.}
    \label{fig:abs}
\end{figure*}

In this work, we introduce LandSegmenter, a {task-specific} FM for LULC mapping, with the goals of: 1) enhancing model flexibility in both input modalities and output categories; and 2) equipping the model with zero-shot capabilities while maintaining its fine-tuning potential. To this end, we build a workflow at three stages as in \cref{fig:abs}. 

First, at the \textbf{input} stage, we curate the \textbf{\ac{LAS} dataset}, which leverages existing \ac{LULC} products as weak supervision to address the scarcity of medium-to-low resolution annotations for model training. {As illustrated in \cref{fig:abs} (a), the exact-to-weak label ratio of LAS is 1:4.} 
It reflects real-world scenarios, in which high-quality annotations are typically limited to high-resolution \ac{RS} imagery, whereas LULC products are predominantly of medium-to-low resolution. LAS also employs more region-level classes to enrich the semantic understanding of Earth surface structures. Through extensive experiments, we demonstrate the effectiveness of weak labels to train segmentation FMs.

Then, to enhance the \textbf{model}’s adaptability for \ac{LULC} mapping, we design \textbf{LandSegmenter} by integrating task-adaptive feature extraction modules with a dynamic fusion strategy. We adopt SAM2's backbone for its robust hierarchical multi-scale spatial feature extraction capability, {complemented with multispectral features from DOFA~\citep{xiong_neural_2024} and detail-enhanced representations from high-frequency components. The additional inputs are aligned with the main feature stream through the \ac{AFM} at intermediate layers.}
Additionally, we replace SAM2’s geometric prompter with the text encoder from GeoRSCLIP~\citep{zhang_rs5m_2024} to boost LandSegmenter’s semantic understanding for flexible and concept-aware output generation. 
{The integration of the text prompter endows LandSegmenter with the zero-shot segmentation capability, which benefits LULC mapping in both training and inference stages. During training, using class names as prompts enables the simultaneous use of multiple heterogeneous datasets. This allows the model to leverage complementary information to improve performance and generalization across various sensors, regions, and spatial resolutions. At inference, users can flexibly generate customized maps under diverse classification needs with a single model, which effectively reduces the effort required to harmonize existing products.}
In this way, LandSegmenter inherits strong generalization ability from existing FMs while gaining explicit semantic understanding for LULC mapping.

Finally, to enhance zero-shot \textbf{inference}, we introduce a \textbf{confidence-guided fusion} strategy to handle semantic omissions. This mechanism uses class-wise confidence scores to guide the fusion of predictions from LandSegmenter and CLIP-style models, thereby improving performance on unseen classes of LandSegmenter. These omitted classes are often object-level entities (e.g., cars) that are absent from standard \ac{LULC} labels but well recognized by CLIP models.

We assess LandSegmenter’s transferability on six precisely annotated \ac{LULC} datasets across different modalities and categories. Results show that LandSegmenter effectively leverages weak supervision to achieve a balance between scalability and precision. We believe our approach offers valuable insights for future research where accurate annotations are scarce but large-scale noisy labels are accessible. 
The main contributions are summarized as follows:
\begin{itemize}[leftmargin=3ex,topsep=2pt,itemsep=0.5ex,partopsep=0ex,parsep=0ex]
\item We propose the first LULC FM termed LandSegmenter, which offers high flexibility in both input and output ends. The model supports zero-shot inference and can also be fine-tuned in downstream tasks.
\item We design a three-stage workflow for constructing LandSegmenter, emphasizing the effective use of large-scale weak supervision to enable scalable FM training, and introducing a class-wise confidence-guided fusion strategy to enhance zero-shot inference.
\item We conduct extensive evaluations across six benchmark LULC datasets with precise annotations. The experimental results demonstrate the effectiveness and generalizability of LandSegmenter under diverse imaging conditions and label granularities.
\end{itemize}

Next, we review related work in \cref{sec:relatedwork}, describe the \ac{LULC} FM construction workflow in \cref{sec:dataset}–\cref{sec:strategy}, present experimental results in \cref{sec:experiments}, and conclude in \cref{sec:conclusion}.

\section{Related Work} \label{sec:relatedwork}
We briefly review recent progress on \ac{FMs}, weakly supervised pretraining, and zero-shot semantic segmentation.

\subsection{Foundation models}
\ac{SSL} plays a crucial role in developing task-agnostic \ac{FMs} from vast unlabeled data~\citep{zhu_foundations_2024, zhao_artificial_2024}. Prominent \ac{SSL} methods include generative \ac{MAE}~\citep{he_masked_2022} and contrastive techniques ~\citep{caron_emerging_2021, chen_improved_2020}. In \ac{EO}, \ac{SSL} is tailored to unique characteristics of RS imagery, such as RingMo’s patch-incomplete \ac{MIM}~\citep{sun_ringmo_2023}, SatMAE’s temporal-spectral embeddings for multispectral data~\citep{cong_satmae_2022,noman_rethinking_2024}, and Scale-MAE’s scale-aware pretraining~\citep{reed_scale-mae_2023}. Recently, multi-modal \ac{SSL} has further advanced cross-modal representation learning~\citep{wang_decur_2023, guo_skysense_2023, fuller_croma_2023, astruc_omnisat_2024}.
Among them, {SkySense++~\citep{wu_semantic-enhanced_2025} focuses on multimodal representation learning across optical and SAR imagery via a semantic-enhanced pretraining strategy.} 
DOFA~\citep{xiong_neural_2024} leverages hypernetworks to generate dynamic patch embedding weights from wavelength, enabling high adaptability across diverse inputs.
While reducing reliance on labeled data, they often require task-specific fine-tuning. The \ac{SAM} models~\citep{kirillov_segment_2023,ravi_sam_2024}, pretrained on millions of natural images or videos and billions of masks, offer a breakthrough as the first segmentation \ac{FMs}~\citep{li_urbansam_2025,zhou_mesam_2024,song_learning_2021,shankar_semantic_2023}. SAM is specific to spatial understanding yet inherently semantic-unaware. Common approaches generate geometric prompts from semantic cues~\citep{chen_rsprompter_2024,wang_sampolybuild_2024} or classify SAM’s segments~\citep{zhang_sam2-path_2024,wang_use_2024}, lacking flexibility and convenience. Besides, SAM is less capable of coping with RS images of diverse modalities and scales due to a lack of relevant training data.

\subsection{Weakly supervised pretraining}
The success of \ac{SAM} relies heavily on vast labeled data. As a cost-efficient alternative, researchers are exploring ``weak'' labels--scalable and affordable, albeit noisy--for model training~\citep{singh_revisiting_2022}. Studies have shown that deep learning models can tolerate some label noise~\citep{zhang_understanding_2021, liu_aio2_2024}. The models pretrained with noisy labels maintain strong feature learning and transferability across tasks like image classification~\citep{mahajan_exploring_2018}, video analysis~\citep{ghadiyaram_large-scale_2019}, and image-text alignment~\citep{jia_scaling_2021}. 
{Several works~\citep{kaiser_learning_2017,maggiori_convolutional_2017,liu_cromss_2025} have explored using pixel-wise weak labels in traditional pretraining paradigms, revealing that shallower layers (closer to the input, typically corresponding to encoders) are less affected by label noise and remain robust after fine-tuning. For instance, CromSS~\citep{liu_cromss_2025} leverages modality-specific encoders within middle and late fusion frameworks during the noisy label pretraining stage, while transferring only the encoders to downstream tasks. In contrast, our work extends the benefits of noisy label pretraining to enhance semantic understanding for zero-shot segmentation. Moreover, unlike CromSS, which relies on separate backbones for different modalities, our model improves multimodal flexibility by handling diverse inputs within a unified framework.}


\subsection{Zero-shot semantic segmentation}
\ac{CLIP} models~\citep{radford_learning_2021} have advanced zero-shot semantic segmentation, also known as \ac{OVSS}, by aligning image and text features to overcome the limitations of closed-set settings~\citep{zhou_image_2024}.
However, \ac{CLIP} is trained at the image level and often struggles to depict details in dense prediction tasks. To alleviate this issue, MaskCLIP leverages value embeddings from CLIP's final layer to improve localization~\citep{zhou_extract_2022}. Others employ self-self attention mechanisms (e.g.\, value-value~\citep{li_closer_2025}, query-query, key-key, or their combinations~\citep{leonardis_clearclip_2024,leonardis_sclip_2024}) to denoise attention maps. \ac{VFM} features have also been integrated to improve CLIP's spatial awareness, either in a training-free ~\citep{leonardis_proxyclip_2024} or training~\citep{shan_open-vocabulary_2024} way. Still, CLIP-based models show reduced sensitivity to RS images. 
{SegEarth-OV} addresses this by introducing a fine-tuned upsampler to recover spatial details~\citep{li_segearth-ov_2024}. Though RS-specific CLIP variants aim to reduce the domain gap with aerial and satellite training data~\citep{zhang_rs5m_2024,liu_remoteclip_2024,wang_skyscript_2024,ye_towards_2024}, they exclusively take RGB as inputs without using multispectral information.

{Beyond CLIP-style models, recent RS \ac{VLMs} such as RemoteSAM~\citep{yao_remotesam_2025}, GeoPixel~\citep{shabbir_geopixel_2025}, and Falcon~\citep{yao_falcon_2025} extend language-guided perception to EO via \ac{VQA} and referring segmentation tasks. However, their heavy reliance on high-resolution RGB imagery and instance-level semantics constrains their effectiveness for dense LULC segmentation involving region-level classes and multispectral data.}

\section{LAS Dataset} \label{sec:dataset}

\begin{figure*}[htp]
  \centering
   \includegraphics[width=1.\linewidth]{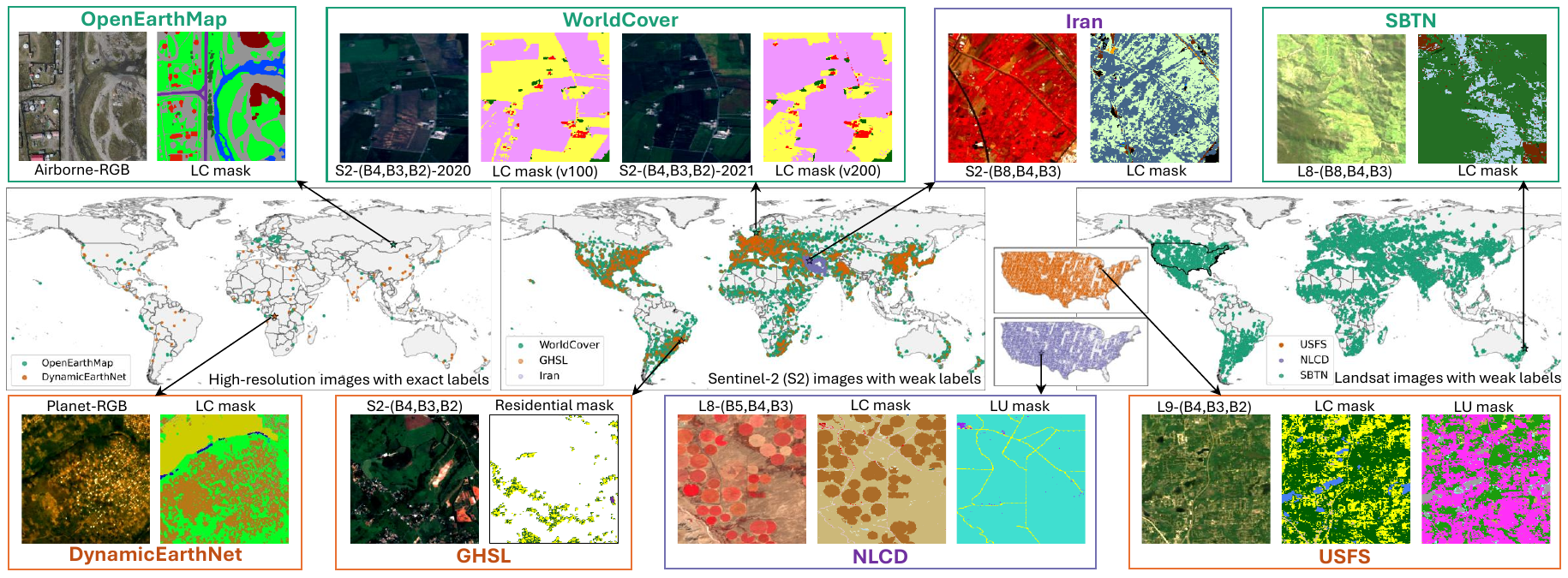}
   \caption{LAS dataset for LandSegmenter training. \textbf{Middle}: geographic distributions of each subset. From left to right, read the distributions of high-resolution, Sentinel-2 (S2), and Landsat-8/9 (L8/9) subsets. \textbf{Top and Bottom}: examples from each subset. Please refer to Appendix for details including the category information and color systems.}
   \label{fig:data}
\end{figure*}

For LULC FM training, we curated the \ac{LAS} dataset, comprising eight subsets from diverse sources, as shown in \cref{fig:data}. Designed to bridge gaps between natural image processing and LULC mapping, LAS addresses: 
\begin{itemize}[leftmargin=3ex,topsep=2pt,itemsep=0.3ex,partopsep=0ex,parsep=0ex]
    \item integration of multispectral RS data beyond RGB;  
    \item adaptation to medium-to-low-resolution RS imagery;
    \item domain knowledge of land surface properties.  
\end{itemize}
As a result, LAS includes $\sim$150k globally distributed sample points ($\sim$311k image patches and $\sim$200k label masks) across eight subsets (see \cref{tab:data}):
\begin{itemize}[leftmargin=3ex,topsep=2pt,itemsep=0.3ex,partopsep=0ex,parsep=0ex]
    \item[1)] high-resolution RGB subset from OpenEarthMap~\citep{xia_openearthmap_2023} (GSD: 0.25–0.5m, patch size: 320);
    \item[2)] RGB-NIR subset from DynamicEarthNet~\citep{toker_dynamicearthnet_2022} (GSD: 3–4m; patch size: 256);
    \item[3)] three Sentinel-2 (S2) subsets (GSD: 10m; 12–13 bands; patch size: 264) following the sampling in~\citet{wang_ssl4eo-s12_2023};
    \item[4)] three Landsat-8/9 (L8/9) subsets (GSD: 30m; 7–11 bands; patch size: 264) following the sampling in~\citet{stewart_ssl4eo-l_2024}.
\end{itemize}
Among them, 1) and 2) are manually annotated, publicly available datasets, while the remaining six pair RS imagery with LULC products downloaded through \ac{GEE} \footnote{https://developers.google.com/earth-engine/datasets/catalog}, resulting in $\sim$80\% weak labels of the whole dataset. This reflects real-world RS data: scarce high-resolution manual annotations versus abundant, imperfect labels often in low resolution. 
The subsets have varied class systems (from coarse to fine, spanning both LC and LU categories), with some designed for specific themes (e.g., residential areas).
\cy{To make these heterogeneous class systems compatible for training, we applied the renaming trick to harmonize them at the dataset/prompt-set level and broaden the text corpus (see the Appendix for the full lists of used class name strings). Within each prompt-defined class system, the class definitions are treated as non-overlapping. More fine-grained ambiguity within a prompt set is not explicitly considered in the current study.}

\begin{table*}[htp]
\centering
\scriptsize
\caption{Details of the eight subsets in LAS. Weakly labeled subsets are named after their paired LULC products. A ``point'' denotes a geospatially unique sampling location. L1C and L2A refer to S2 processing levels, corresponding to Top-of-Atmosphere (TOA) reflectance and Surface Reflectance (SR) images, respectively. ``Res'' and ``Imp'' are short for residential and impervious.}
\setlength{\extrarowheight}{0.3mm}
\setlength\tabcolsep{6pt}
\begin{tabular}{>{\centering}m{0.6cm}|>{\centering}m{1.1cm}|>{\centering}m{.6cm}|>{\centering}m{.8cm}|>{\centering}m{1.1cm}|>{\centering}m{1.1cm}|>{\centering}m{.6cm}|>{\centering}m{.6cm}|>{\centering}m{.65cm}|m{5.8cm}}
\toprule
\textbf{Label}           & \textbf{Name}                  & \hspace{-.25cm}\textbf{\#point}                & \textbf{Scope}                & \textbf{Sensor}                            & \textbf{GSD}           & \hspace{-.2cm}\textbf{\#band}                        & \textbf{Year}                  & \textbf{\#class}  & \makecell{\textbf{Notes}}       \\
\midrule
\multirow{3}{*}{Exact} & \makecell{Open\\EarthMap}
& 25.0k                  & Global               & Multiple & 0.25-0.5m            & 3                              & -                     & 8       &  \makecell{Big tiles were cropped to small patches (320x320).}      \\
\cline{2-10}
                       & Dynamic EarthNet
                       & 4.8k                   & Global               & Planet                            & 3-4m                 & 4                              & 2018- 2019             & 7        &  Big tiles were cropped to small patches (256x256). Each point has the data from 4 seasons.      \\
\hline
\multirow{12}{*}{Weak}  & Iran                  & 5.5k                   & Iran                 & S2(L1C)                          & 10-60m               & 13                             & 2017                  & 10         &  \makecell{All bands were upsampled to 10m.}    \\
\cline{2-10}
                       & GHSL                  & 9.3k                   & Global               & S2(L1C)                          & 10-60m               & 13                             & 2018                  & Res:3  & \makecell{All bands were upsampled to 10m.} \\
                       \cline{2-10}
                       & World Cover v100/200   & 44.0k                  & Global               & S2(L1C /L2A)                      & 10-60m               & 12                             & 2020-2021             & 11              & Each point corresponds to two data triples: (L1C, L2A, LC-v100) and (L1C, L2A, LC-v200) from years 2020 and 2021. The L1C-B10 (cirrus) band is discarded to ensure band consistency with L2A. \\
                       \cline{2-10}
                       & \multirow{2}{*}{NLCD} & \multirow{2}{*}{18.7k} & \multirow{2}{*}{USA} & L8             & \multirow{2}{*}{30m} & \multirow{2}{*}{7/11} & \multirow{2}{*}{2019} & LC:16         & Each point corresponds to a data quadruple:  \\
                       &                       &                        &                      &    (SR/TOA)                   &                      &                                &                       & Imp:3 &  (SR, TOA, LC, Imp).\\
                       \cline{2-10}
                       & \multirow{2}{*}{USFS} & \multirow{2}{*}{18.7k} & \multirow{2}{*}{USA} & L9              &  \multirow{2}{*}{30m} & \multirow{2}{*}{7/11}               & \multirow{2}{*}{2023} & LC:12          & Each point corresponds to a data quadruple: \\
                       &                       &                        &                      &    (SR/TOA)                   &                      &                                &                       & LU:5 &  (SR, TOA, LC, LU).\\
                       \cline{2-10}
                       & \multirow{2}{*}{SBTN}                  & \multirow{2}{*}{22.7k}                  & \multirow{2}{*}{Global}               & L8                               &  \multirow{2}{*}{30m}              &   \multirow{2}{*}{7/11}                          & \multirow{2}{*}{2020}                  & \multirow{2}{*}{11}              & Each point corresponds to a data triple: \\
                       &                                        &                                         &                                       & (SR/TOA)  & & & & &
                       (SR, TOA, LULC). \\
\midrule
\multicolumn{2}{c|}{\textbf{Total}}                     & \multicolumn{1}{l|}{148.6k} & Global & Multiple &  0.25-30m & 3-13 & \hspace{-.2cm}\;-2023 & 3-16 & Images of different processing levels serve as a data augmentation strategy during training. \\
\bottomrule
\end{tabular}
\label{tab:data}
\end{table*}

\section{LandSegmenter Model} \label{sec:method}
We introduce the architecture of LandSegmenter in \cref{sec:method:arch}, followed by its training details in \cref{sec:method:loss}.

\begin{figure*}[htp]
  \centering
   \includegraphics[width=1.\linewidth]{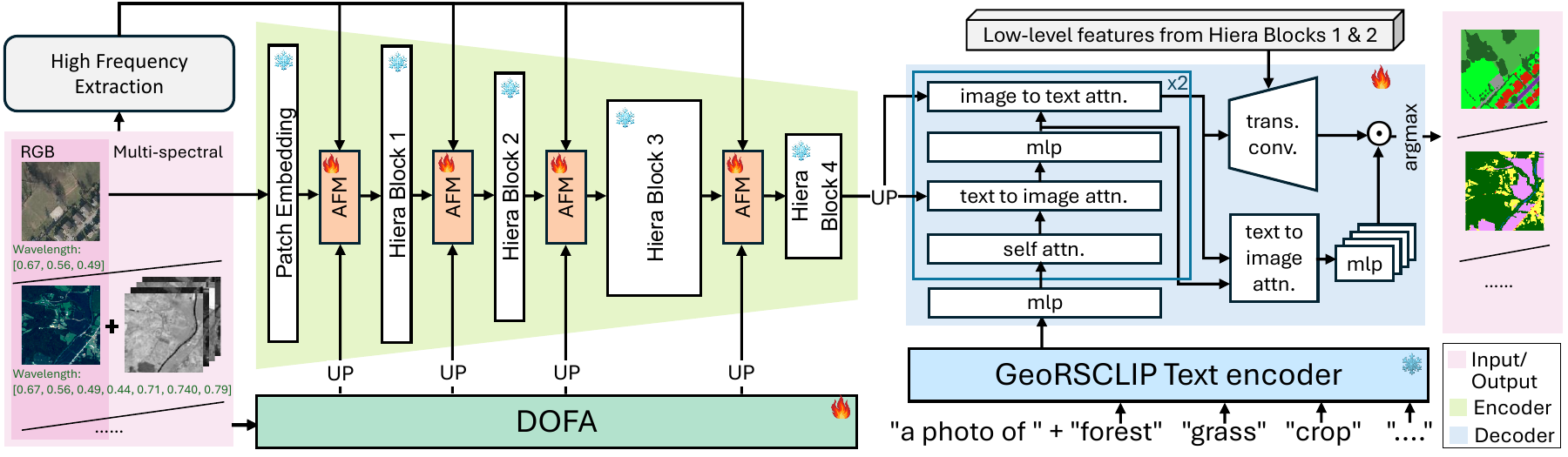}
   \caption{Architecture of LandSegmenter, where the attention-based fusion module (AFM) is depicted per block to indicate the consistent additional input at every stage, with its layer-wise implementation detailed in \cref{fig:meth:afm}. {The embeddings sent to the decoder are the summation of the outputs from Blocks 4 (upsampled) and 3. For simplicity, we omit this operator in the figure.}}
   \label{fig:meth:architect}
\end{figure*}

\subsection{Architecture} \label{sec:method:arch}
LandSegmenter has three key components: an RS-imagery-adaptive visual encoder, an LULC class name text prompt encoder, and a vision-text collaborative decoder.

\paragraph{\textbf{Encoder}}

\begin{figure}
  \centering   \includegraphics[width=1\linewidth]{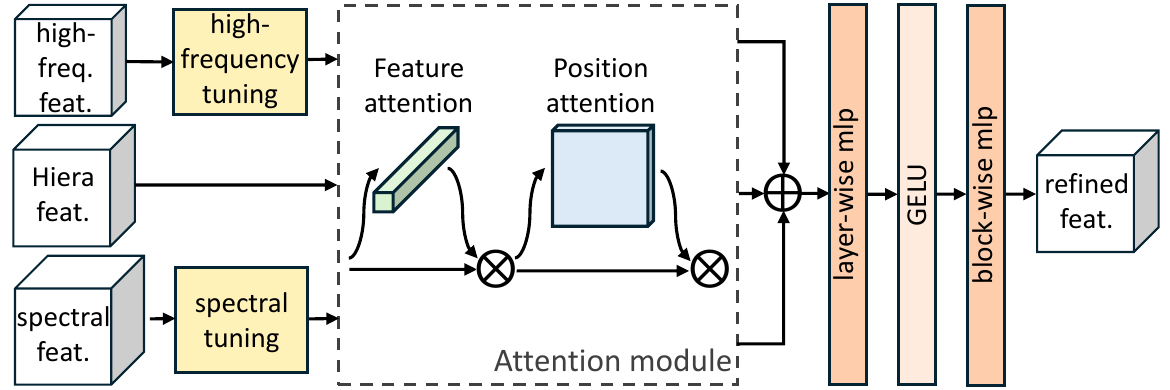}
   \caption{Attention-based fusion module (AFM), where the attention modules share the same architecture yet are individually optimized for each input.}
   \label{fig:meth:afm}
\end{figure}

LandSegmenter’s encoder adopts SAM2’s Hiera backbone as its core structure in a hierarchical fashion. Drawing inspiration from \citet{chen_sam2-adapter_2024} and \citet{ferrari_cbam_2018}, we incorporate an attention-based adapter to enhance multispectral RS image processing. As shown in \cref{fig:meth:architect}, \ac{AFM} is inserted to align the main-steam features and the two specialized components as follows:
\begin{itemize}[leftmargin=3ex,topsep=2pt,itemsep=0.5ex,partopsep=0ex,parsep=0ex]
    \item A \textbf{High-Frequency (HF) Extractor} that strengthens low-level features and address detail loss;
    \item The \textbf{DOFA Model}~\citep{xiong_neural_2024} that processes multispectral imagery to enrich spectral information. 
\end{itemize}
For HF component extraction, we use the Fast Fourier Transform (\texttt{fft}) and its inverse (\texttt{ifft}) following \citep{liu_explicit_2023}. Let $\mathbf{Z}=\text{\texttt{fft}}(\mathbf{I}^{(c)})$ be the frequency component of the $c$th channel of image $\mathbf{I} \in \mathbb{R}^{H \times W \times C}$. The HF feature $\mathbf{I}_{FH}^{(c)}$ is:
\begin{equation} \label{eq:hf}
\small
    \mathbf{I}_{FH}^{(c)}=\text{\texttt{ifft}}(\mathbf{Z}\cdot\mathbf{M}(\tau)),
\end{equation}
where $\mathbf{M}(\tau)$ is a binary mask eliminating low-frequency coefficients from the image center given the mask ratio $\tau$. We apply \cref{eq:hf} to each band and standardize the number of HF component inputs across modalities to six as, 
\begin{equation}
\small
    \mathbf{I}_{FH}=[\mathbf{I}_{FH}^{(R)}, \mathbf{I}_{FH}^{(G)}, \mathbf{I}_{FH}^{(B)}, \;\min_{c}(\mathbf{I}_{FH}^{(c)}), \;\max_{c}(\mathbf{I}_{FH}^{(c)}), \;\overline{\mathbf{I}_{FH}^{(c)}}]. 
\end{equation}
For spectral information extraction, DOFA utilizes a hypernetwork to dynamically generate band-specific patch embedding kernels given the central wavelength, leading to modality-tailored features~\citep{xiong_neural_2024}. To balance computational efficiency, we adopt DOFA-base to extract the spectral input $\mathbf{I}_{spe}^{b}$ for Hiera block {$b=\{1,2,3,4\}$}:
\begin{equation}
\label{eq:dofa}
\small
    \mathbf{I}_{spe}^{(b)}=\text{DOFA}_{l_b}(\mathbf{I}, \mathbf{w}),
\end{equation}
where {$l_b=\{1, 4, 9, 11\}$} denotes the DOFA output layer indices for Hiera block $b$, $\mathbf{w}$ is $\mathbf{I}$'s central wavelength vector.

In \ac{AFM}, we follow \citet{ferrari_cbam_2018} to entangle the three kinds of features with the feature and position attention as demonstrated in \cref{fig:meth:afm}. Let $\mathbf{E}_i$ be the Hiera output from the $i$th layer. The refined features by AFM are,
\begin{equation}
\small
    \mathbf{I}_{ref}^{(b,i)} = \texttt{MLP}_{b}(\texttt{GELU}(\texttt{MLP}_{i}(\widetilde{\mathbf{E}}_i+\widetilde{\mathbf{I}}_{HF}+\widetilde{\mathbf{I}}_{spe}^{(b)}))),
\end{equation}
where $\texttt{MLP}_{b}$ and $\texttt{MLP}_{i}$ are block-wise and layer-wise multi-layer perceptrons,  
\{$\widetilde{\mathbf{E}}_i$, $\widetilde{\mathbf{I}}_{HF}$, $\widetilde{\mathbf{I}}_{spe}^{(b)}$\} are attention-enhanced features derived separately from $\{\mathbf{E}_i, \mathbf{I}_{HF}, \mathbf{I}_{spe}^{(b)}\}$ with the operation $\widetilde{\mathbf{F}} = \mathbf{F} \otimes \texttt{a}_{f}(\mathbf{F}) \otimes \texttt{a}_{p}(\mathbf{F} \otimes \texttt{a}_{f}(\mathbf{F}))$, and $\texttt{a}_{f}$ and $\texttt{a}_{p}$ represent feature and position attention modules, which can be formulated as follows:
\begin{equation}
\begin{split}
    \texttt{a}_f(\mathbf{F}) =&  \sigma (\texttt{MLP}(\mathbf{F}_{AP}^{s})+\texttt{MLP}(\mathbf{F}_{MP}^{s})), \\
    \widetilde{\mathbf{F}}_f =& \mathbf{F} \otimes \texttt{a}_f(\mathbf{F}), \\
    \texttt{a}_p(\widetilde{\mathbf{F}}_f) =& \text{conv}([\text{max}_c(\widetilde{\mathbf{F}}_f);\text{mean}_c(\widetilde{\mathbf{F}}_f)]),
\end{split}
\end{equation}
where $\mathbf{F}_{AP}^{s},\mathbf{F}_{MP}^{s}\in\mathbb{R}^{B \times C}$ are the features obtained with spatial average and max pooling from $\mathbf{F}$, $\sigma$ is the sigmoid function, $\text{max}_c$ and $\text{mean}_c$ are operated along the channel dimension (dim=1), $\otimes$ is the element-wise multiplication, and $\text{conv}$ is with a kernel size of 7. 
For $\mathbf{I}_{FH}$ and $\mathbf{I}_{spe}^{(b)}$, a tuning block, comprising three linear layers interleaved with GELU activations, aligns their feature dimensions with $\mathbf{E}_i$ before input to the attention module.

\paragraph{\textbf{Prompter}}

To enhance semantic awareness for \ac{LULC} mapping, we replace the original geometric prompter with a text encoder from GeoRSCLIP \citep{zhang_rs5m_2024}. GeoRSCLIP was pretrained on a large corpus of RS image-text pairs with geolocation-informed descriptions, making it better-suited for \ac{LULC} tasks than other vanilla and RS-based CLIP models (see \cref{sec:exp:ablation:arch}). \cy{LandSegmenter directly exploits the fixed pretrained semantic space provided by GeoRSCLIP with the text encoder kept frozen during training.} We freeze the text encoder due to the relatively limited text corpus compared to the abundance of images in LAS. \cy{Accordingly, semantic harmonization across heterogeneous datasets is achieved at the prompt level via the renaming trick. Specifically, we reformulate and expand dataset-specific class names into more descriptive, semantically standardized prompts. Therefore, they can be more consistently aligned with this fixed semantic space. Further details of the renaming trick are provided in the Appendix.}

\paragraph{\textbf{Decoder}}

As illustrated in \cref{fig:meth:architect}, the trainable decoder integrates image embeddings with prompter text embeddings to generate the final outputs. \cy{It takes as input the fused visual embeddings \(\mathbf{F}_v \in \mathbb{R}^{H \times W \times C}\) and the projected prompt text embeddings \(\mathbf{F}_t \in \mathbb{R}^{K \times C}\), where \(K\) is the number of class prompts. The interaction between text and visual features is performed by two stacked Two-Way Transformer blocks. Denoting the generic cross-attention operator as}
\begin{equation}    
\cy{\mathrm{CA}(\mathbf{Q},\mathbf{K},\mathbf{V})=\mathrm{Softmax}\!\left(\frac{\mathbf{Q}\mathbf{K}^{\top}}{\sqrt{d}}\right)\mathbf{V},}
\end{equation}
\cy{each block first applies self-attention to the text tokens, followed by a text-to-image cross-attention layer with \(\mathbf{Q}=\mathbf{F}_t\) and \(\mathbf{K}=\mathbf{V}=\mathbf{F}_v\), an MLP on the updated text tokens, and finally an image-to-text cross-attention layer with \(\mathbf{Q}=\mathbf{F}_v\) and \(\mathbf{K}=\mathbf{V}=\mathbf{F}_t\). In this way, the text branch is enhanced with image-aware spatial evidence. The image branch is enriched with class-aware semantic information from the class name prompts.}

\cy{The enriched visual features are progressively upsampled by a two-layer transpose-convolution head. With shallow feature injection, the two stages can be written as}
\begin{equation} 
\begin{aligned}
\cy{\mathbf{U}_1=\sigma\!\left(\mathrm{LN}\!\left(\mathrm{ConvTrans}_1(\mathbf{F}_v)\right)+\mathbf{S}_2\right),} \\
\cy{\mathbf{U}_2=\sigma\!\left(\mathrm{ConvTrans}_2(\mathbf{U}_1)+\mathbf{S}_1\right),}
\end{aligned}
\end{equation}
\cy{where \(\mathbf{S}_1\) and \(\mathbf{S}_2\) denote the projected low-level features from shallow Hiera blocks, \(\mathrm{LN}(\cdot)\) denotes layer normalization, and \(\sigma(\cdot)\) denotes the activation function. This design restores fine spatial details while simultaneously preserving the semantic information carried by the high-level fused features.}

\cy{After the stacked Two-Way Transformer blocks, a final text-to-image attention layer is applied to further update the text tokens, which are then fed into MLP hypernetworks to generate class-specific prediction weights. These weights are applied to the upsampled dense features to produce the pre-softmax class score maps, enabling flexible LULC segmentation across diverse categories.}

\subsection{Training} \label{sec:method:loss}

We employ a combined loss function of CrossEntropy (CE) and Dice~\citep{jadon_survey_2020} for training. Predictions are generated class-wise before the softmax layer. Thus, we incorporate both multi-class and binary losses in practice:
\begin{equation}
\begin{aligned} 
\small
    L=&\texttt{CE}(\widetilde{\mathbf{Y}},\mathbf{Y})+\texttt{Dice}(\widetilde{\mathbf{Y}},\mathbf{Y}) \\
    &+\sum_{k=1..K}\big(\texttt{BCE}(\widetilde{\mathbf{Y}}_k,\mathbf{Y}_k)+\texttt{BDice}(\widetilde{\mathbf{Y}_k},\mathbf{Y}_k)\big),
\end{aligned}
\end{equation}
where $\widetilde{\mathbf{Y}}$ and $\mathbf{Y}$ denote predicted and ground-truth (GT) label masks, and $k$ is the class index.

Taking into consideration the computational cost and pretrained nature of SAM2 and DOFA, we freeze the Hiera backbone and apply a reduced learning rate to DOFA (0.1 of the others) to preserve their foundational capabilities.

The six S2 and L8/9 subsets of the LAS dataset introduce label noise, which can bias the training process. As noted by~\citet{liu_task_2024}, such semantic noise disproportionately affects the deeper, semantically richer layers. To mitigate overfitting to label noise, we employ an auxiliary decoder during training. Specifically, we adopt a Siamese-like architecture in which only the decoder is duplicated. The encoder is shared and processes all input data, while the two decoders are trained in an alternating fashion: the main decoder handles odd-numbered batches (1, 3, ...), and the auxiliary decoder processes even-numbered batches (2, 4, ...). This simple yet effective strategy has been widely used for handling label noise~\citep{Ouali2020Decoding}. During inference or transfer, only the main decoder is retained.

\begin{figure}
    \centering
\includegraphics[width=1.\linewidth]{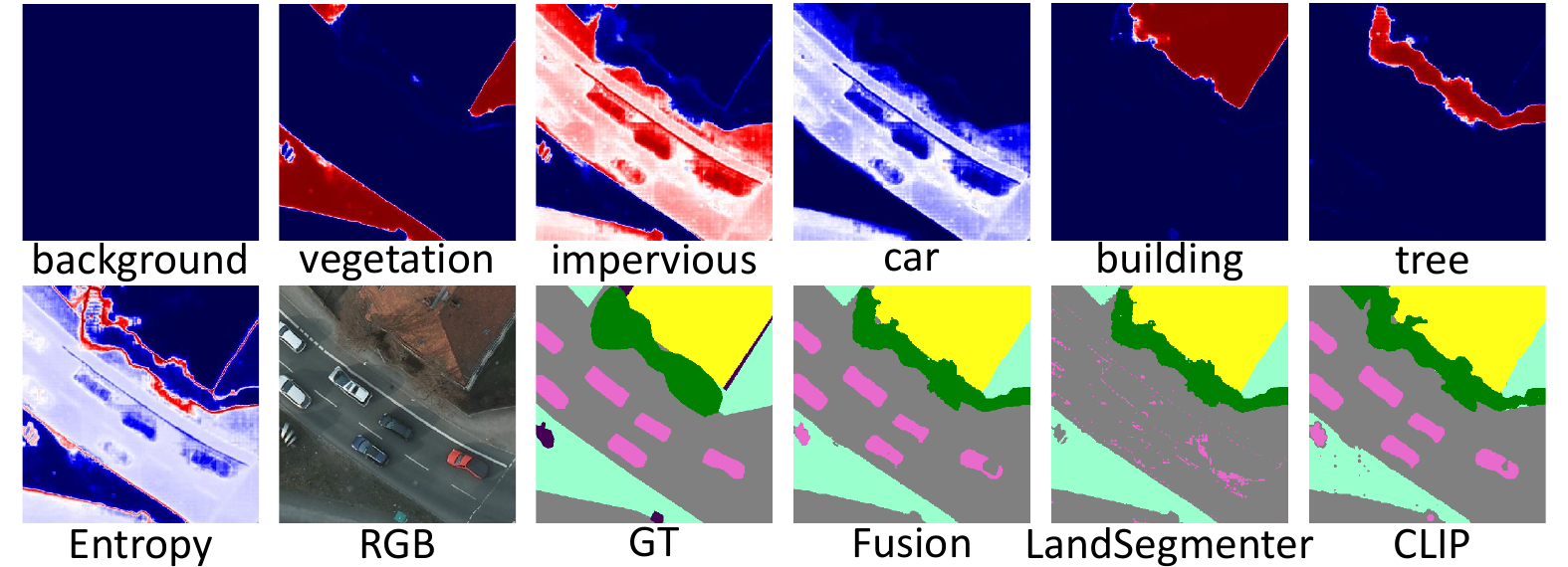}
    \caption{An example from Potsdam where car is absent in the LAS dataset. \textbf{Top}: class-wise confidence maps from softmax outputs. \textbf{Bottom}: pixel-wise uncertainty map (entropy of probability vectors); RGB image; \ac{GT} mask; prediction by the confidence-guided fusion strategy (Fusion); prediction by LandSegmenter; prediction by ProxyCLIP with the features refined with LandSegmenter's embeddings (CLIP). Confidence and uncertainty values range from 0 (blue) to 1 (red). The class scheme of \ac{GT} and predictions is the same as that in \cref{tab:exp:potsdamclass}.}
    \label{fig:strategy:confidence}
\end{figure}

\section{Confidence-guided Fusion for Zero-shot Inference}\label{sec:strategy}

To boost LandSegmenter’s zero-shot performance on unseen classes beyond LAS, we introduce a class-wise confidence-guided fusion strategy (see \cref{fig:abs} (c)). As demonstrated in \cref{fig:strategy:confidence}, we find that LandSegmenter often misclassifies unseen objects (e.g., car as impervious) with low entropy-based uncertainty. At the same time, the class-wise confidence maps for unseen classes remain consistently lower than those for the in-distribution classes of LAS. Motivated by this, \cy{as illustrated in \cref{fig:abs} (c),} we use the maximum confidence score $C^{k}=\max(\mathbf{P}^{k})$ from each class-wise predicted probability map $\mathbf{P}^{k}$ as a fusion indicator: given a predefined confidence threshold $C_t$, we treat $\mathbf{P}^{k}$ with $C^{k} > C_t$ as confident (in-distribution), and those with $C \leq C_t$ as of low confidence (out-of-distribution or irrelevant classes). In low-confidence cases of LandSegmenter, we prioritize CLIP’s prediction over LandSegmenter’s using a weighted fusion ratio of 3:1 if CLIP shows high certainty. Otherwise, we average predictions from both models with a balanced 2:2 ratio. Formally, let $\mathbf{P}_{land}^{k}$ and $\mathbf{P}_{clip}^{k}$ denote the predicted probability maps for class $k$ from LandSegmenter and CLIP, respectively. The fusion process can be formulated as follows,
\begin{equation}
\small
    \mathbf{P}_{f}^{k} = 
    \begin{cases}
      1\times\mathbf{P}_{land}^{k} + 3\times \mathbf{P}_{clip}^{k} & \text{if} \; C_{land}^{k} \leq C_t \; \text{and} \; C_{clip}^{k} > C_t,\\
      2\times\mathbf{P}_{land}^{k} + 2\times \mathbf{P}_{clip}^{k} & \text{otherwise}.
    \end{cases}
\end{equation}
We further enhance the CLIP-based predictions $\mathbf{P}_{clip}^{k}$ by incorporating LandSegmenter’s encoder features via ProxyCLIP~\citep{leonardis_proxyclip_2024} before the fusion step. Briefly, ProxyCLIP employs \ac{VFM} features—here, from LandSegmenter—as queries and keys to compute the attention map, and use the CLIP features as values to produce refined predictions. For simplicity, we use PC to represent ProxyCLIP in the following. As shown at the bottom of~\cref{fig:strategy:confidence}, the proposed fusion strategy enables the recovery of unseen classes (e.g., car) from CLIP, while retaining confident predictions from LandSegmenter (e.g., low vegetation at the top).

\section{Experiments} \label{sec:experiments}

\begin{table*}[htp]
\footnotesize
\centering
\caption{Zero-shot segmentation performance (mIoU/OA) on six test datasets.
The best and second-best results are highlighted in \textbf{bold} and \underline{underlined}.}
\setlength{\extrarowheight}{0.3mm}
\setlength\tabcolsep{3.3pt}
\begin{tabular}{c|cccccc|c}
\toprule
  \textbf{Dataset}        & \textbf{Potsdam} & \textbf{LoveDA} & \textbf{NYC}   & \textbf{DW}    & \textbf{OSM}   & \textbf{MultiSenGe} & \multirow{4}{*}{\textbf{Average}} \\
\cline{1-7}
\textbf{GSD} & 0.05m  & 0.3m  &  1m  &  10m  &  10m  & 10m  &  \\
\cline{1-7}
\textbf{Image (\#band)}  & RGB (3)  & RGB (3)  & RGB (3)   & S2 (13)  & S2 (13) & S2 (10) &  \\
\cline{1-7}
\textbf{\#class} & 6  & 7  &  8  &  9  &  13  & 14  &   \\
\midrule
vanilla CLIP~\citep{radford_learning_2021} & 32.21/57.45 & 27.59/46.63 & 18.35/34.83 & 23.99/52.55 & 10.54/48.18 & 7.43/50.14 & 20.02/48.30 \\
MaskCLIP~\citep{zhou_extract_2022}     & 34.69/55.83 & 27.80/42.43 & 19.46/38.55 & 19.39/43.96 &  9.56/41.02 & 9.50/45.84 & 20.07/44.61 \\
SCLIP~\citep{leonardis_sclip_2024}        & 40.06/63.91 & 33.33/51.24 & 23.83/43.89 & 25.91/54.00 & 11.60/47.41 & 9.98/56.61 & 24.12/52.84 \\
ClearCLIP~\citep{leonardis_clearclip_2024}    & 39.38/63.56 & 32.79/52.30 & 23.33/41.94 & 25.85/54.42 & 11.90/48.50 & 9.91/55.55 & 23.86/52.71 \\
SegEarth-OV~\citep{li_segearth-ov_2024}     & 45.28/68.11 & 36.91/54.40 & 25.12/46.35 & 28.23/55.74 & 13.49/52.87 & 12.38/60.47 & 26.90/56.32 \\
PC (w DINO)~\citep{leonardis_proxyclip_2024}                           & 43.08/66.61            & 26.03/38.25           & 20.65/36.88          & 37.50/62.57          & 22.52/54.97          & 12.79/64.06               & 27.61/53.25            \\
PC (w SAM2)~\citep{leonardis_proxyclip_2024}                           & 41.90/64.32            & 25.54/37.70           & 20.00/34.74          & 35.76/60.88          & 20.60/53.66          & 11.55/62.01               & 25.55/52.44            \\
\hline
RemoteCLIP~\citep{liu_remoteclip_2024}     & 21.38/40.24 & 37.22/56.63 & 24.05/45.79 & 23.95/48.92 &  7.32/29.54 &  8.53/41.04 & 20.41/43.69 \\
GeoRSCLIP~\citep{zhang_rs5m_2024}      & 39.78/66.23 & 31.56/50.03 & 27.38/48.99 & 27.58/57.13 & 12.58/56.15 & 13.99/59.86 & 25.48/56.40 \\
SkyCLIP~\citep{wang_skyscript_2024}        & 40.44/67.53 & 32.14/47.87 & 23.60/44.91 & 23.96/51.46 &  8.65/35.56 &  9.07/51.29 & 22.98/49.77  \\
\hline
RemoteSAM~\citep{yao_remotesam_2025}        &  \textbf{64.05/77.02}  &  20.44/39.35  &  7.82/16.69  &  6.03/17.61  &  1.31/2.64  &  1.77/0.43  &  16.90/25.62  \\
GeoPixel~\citep{shabbir_geopixel_2025}      &  24.19/44.76  &  19.21/31.75  &  8.86/19.68  &  21.05/39.95  &  12.98/38.72  &  8.95/3.76  &  15.87/29.77  \\
\midrule
PC (w LandSegmenter) & 43.65/68.50            & 27.47/39.81           & 22.16/39.01          & 40.00/65.36          & 24.20/58.02          & 13.99/\textbf{67.21}               & 28.58/56.32            \\
LandSegmenter               & 41.53/72.21            & {\ul 40.40}/{\ul 58.97}     & {\ul 31.44}/\textbf{53.69}          & {\ul 44.08}/{\ul 67.37}          & {\ul 29.35}/{\ul 71.93}          & {\ul 18.07}/62.89               & {\ul 34.15}/{\ul 64.51}            \\
Confidence-guided Fusion             & {\ul 49.73}/{\ul 75.43}   & \textbf{40.87}/\textbf{59.15}  & \textbf{33.34}/{\ul 53.54}    & \textbf{46.06}/\textbf{69.35}    & \textbf{30.69}/\textbf{74.03} & \textbf{18.92}/{\ul 66.90}               & \textbf{36.60}/\textbf{66.40}      \\
\bottomrule
\end{tabular}
\label{tab:exp:ov}
\end{table*}

We trained LandSegmenter on the LAS dataset for 50 epochs using the AdamW optimizer~\citep{loshchilov_decoupled_2018}, with an initial learning rate of $1e-4$ decaying to $1e-6$ via a cosine scheduler. Inputs were randomly cropped to 256$\times$256, with random flipping and rotation for augmentation, and then resized to the required sizes by Hiera and DOFA. Batch size is set to 12 on each GPU. Training on 4 NVIDIA H100 GPUs took $\sim$44 hours. We evaluated LandSegmenter’s performance through zero-shot and fine-tuning experiments on six LULC datasets: 
\begin{itemize}[leftmargin=3ex,topsep=2pt,itemsep=0.5ex,partopsep=0ex,parsep=0ex]
    \item \textbf{Potsdam}\footnote{https://www.isprs.org/education/benchmarks/UrbanSemLab/} is a very-high-resolution dataset of 5cm created for urban semantic segmentation. It includes a training split with 24 big tiles and a validation split with 14 big tiles. We crop them to small patches of 512$\times$512, resulting in 2904 and 1694 training and test patches in our experiments. We utilize RGB images as inputs and generate segmentation maps of 6 classes as in \cref{tab:exp:potsdamclass}.
    \item \textbf{LoveDA}~\citep{wang_loveda_2022} is constructed with 0.3m RGB images obtained from the \ac{GEE} platform over three Chinese cities, paired with 7-class label masks as in \cref{fig:cmap:loveda}. We crop the initial 1024$\times$1024 tiles to 512$\times$512, leading to 9718 and 6505 for training and testing after removing no-data patches.
    \item \textbf{NYC}~\citep{albrecht_monitoring_2022} is from the publicly available data for the area of New York City (NYC). We utilize NAIP\footnote{https://naip-usdaonline.hub.arcgis.com/}'s RGB images as inputs. The \ac{GT} masks are provided by the NYC agencies generated based on the 2017 NYC LiDAR survey and other supplementary information with 8 classes as in \cref{fig:cmap:nyc}. We have 6000 training patches and 4000 test patches of 256$\times$256 pixels.
    \item \textbf{DW}~\citep{liu_cromss_2025} collects the image-label pairs from the training and test sets of the Google Dynamic World (DW) project \citep{brown_dynamic_2022}. The input images are Sentinel-2 L1C images fetched according to the date and coordinates of the label masks. It contains 9 basic LC classes as in \cref{fig:cmap:dw}. The label masks are not densely annotated, leaving uncertain parts unlabeled. We crop them to 256$\times$256, leading to 14163 for training and 1359 for testing after removing no-data patches.
    \item \textbf{OSM}~\citep{liu_cromss_2025} is an extension of DW, with labels from OpenStreetMap (OSM)\footnote{https://www.openstreetmap.org/}. The labels are cross-checked with those from DW, plus some manual checks to ensure the quality. The labels are even sparser than those of DW due to the volunteered geographic information nature of OSM. Nevertheless, the class categories are finer with many land use classes as in \cref{fig:cmap:osm}. We have 4821 training and 1428 testing patches of 256$\times$256. 
    \item \textbf{MultiSenGe}~\citep{wenger_multimodal_2023} is constructed from 14 Sentinel-2 L2A tiles over the GrandEst region in France. The reference data is from the Land Use Land Cover Database (BDOCGE2) provided by French administrators. The dataset is with 10 bands after excluding 2 low-resolution bands (B1, B10). We randomly split 8157 patches of 256$\times$256 into 4157 for training and 4000 for testing. As shown in \cref{fig:cmap:multisenge}, its 14 classes have many land use ones, making it challenging for LULC mapping.
\end{itemize}
These datasets were chosen for varying modalities and class definitions, and more importantly, label reliability. The confidence threshold $C_t$ for confidence-guided fusion is empirically set to 0.6, with its sensitivity analyzed in \cref{sec:exp:hyper}. The name text prompts used in our experiments are listed in the Appendix. 
For fine-tuning, we use subsets (0.1 and 0.3) of the training set to evaluate LandSegmenter’s transfer learning capability. We set the initial learning rates for SAM2-related models and other comparison methods to $5e-4$ and $1e-4$, respectively, with cross-validation. The cosine scheduler is used in all the cases to adaptively decay the learning rate to $1e-6$. The batch size is 15. The total number of fine-tuning epochs is fixed as 30. We use random flipping and rotation with a rate of 0.5 and 0.2 as data augmentation. One NVIDIA H100 GPU is used for fine-tuning.


\subsection{Main results}
\subsubsection{Zero-shot}

\begin{table}[htp]
\centering
\footnotesize
\caption{Class-wise zero-shot results on Potsdam with bg, veg, imp, bd, LandSeg being background, low vegetation, impervious, building, and LandSegmenter.}
\setlength{\extrarowheight}{0.mm}
\setlength\tabcolsep{4pt}
\begin{tabular}{c|cccccc}
\toprule
\multirow{2}{*}{\textbf{Potsdam}}                   & \multicolumn{6}{c}{\textbf{IoU (\%)}}\\
\cline{2-7}
   & \textcolor{white}{\textbf{bg}}\cellcolor[HTML]{440154} & \textbf{veg}\cellcolor[HTML]{99FFCC} & \textbf{imp}\cellcolor[HTML]{808080} & \textbf{car}\cellcolor[HTML]{E86ACD}   & \textbf{bd}\cellcolor[HTML]{FFFF19} & \textbf{tree}\cellcolor[HTML]{008000}                           \\
\midrule
{SegEarth-OV}                                   & {\ul 14.07}      & 50.86            & 59.81                & 48.37 & 57.27    & 41.33  \\
PC (w DINO)                         & {12.23}      & 49.80            & 59.58                & 51.29 & 55.16    & 30.45  \\
PC (w SAM2)                         & 9.58       & 47.38            & 57.01                & {\ul 59.84} & 50.46    & 27.10  \\
\hline
RemoteCLIP & 9.06&	6.00&	2.35&	11.74&	69.82&	29.34 \\
GeoRSCLIP & 3.76&	49.24&	51.11&	19.46&	63.52&	\textbf{51.59} \\
SkyCLIP & 2.12&	50.98&	57.39&	31.56&	59.69&	40.88 \\
\hline
{RemoteSAM}  & \textbf{59.85} & 33.98 & {\ul 68.55} & \textbf{79.68} & \textbf{92.70} & {\ul 49.54} \\
{GeoPixel}   & 7.34 & 29.23 & 41.73 & 14.99 & 47.65 & 4.20 \\
\midrule
 PC (w LandSeg) & 10.41      & 52.22            & 62.18                & 48.28 & 59.99    & 28.85 \\
LandSeg               & 10.18       & {\ul 53.68}            & 63.10                & 1.23  & {80.57}    & 40.40  \\
Fusion            & 11.16      & \textbf{55.46}            & \textbf{68.63}                & 39.49 & {\ul 81.14}    & 42.50  \\
\bottomrule
\end{tabular}
\label{tab:exp:potsdamclass}
\end{table}

We compare LandSegmenter against state-of-the-art \ac{OVSS} methods and report their mIoU and Overall Accuracy (OA) scores in \cref{tab:exp:ov}. Specifically, {we evaluate six CLIP-based methods (vanilla CLIP~\citep{radford_learning_2021}, MaskCLIP~\citep{zhou_extract_2022}, SCLIP~\citep{leonardis_sclip_2024}, ClearCLIP~\citep{leonardis_clearclip_2024}, SegEarth-OV~\citep{li_segearth-ov_2024}, and ProxyCLIP (PC)~\citep{leonardis_proxyclip_2024}), three RS-specific CLIP variants (RemoteCLIP\citep{liu_remoteclip_2024}, GeoRSCLIP~\citep{zhang_rs5m_2024}, and SkyCLIP~\citep{wang_skyscript_2024}), as well as two RS VLMs (RemoteSAM~\citep{yao_remotesam_2025} and GeoPixel~\citep{shabbir_geopixel_2025}). For RS-specific CLIP variants, we follow~\citet{li_segearth-ov_2024} and apply FeatUp to enhance detail preservation.}
{Note that we use ``zero-shot'' to refer to applying models to unseen datasets. In this case, models handle both seen and unseen categories during evaluation. This setup reflects real-world LULC scenarios, where datasets often exhibit partial class overlap.}
As shown in \cref{tab:exp:ov}, LandSegmenter outperforms other considered methods, except on the Potsdam dataset with degraded performance on the unseen car class (see \cref{tab:exp:potsdamclass}). 
{Here, RemoteSAM performs well on the very-high-resolution Potsdam dataset. RemoteSAM's training data include Potsdam. Its referring-based training paradigm puts more focus on instance-level targets. Thus, RemoteSAM achieves very high accuracy on object-level categories such as car and building, but performs worse on region-level classes such as low vegetation and impervious surfaces. LandSegmenter shows strong performance, especially on the three low-resolution multispectral datasets, which highlights LandSegmenter's robustness in challenging RS scenarios.}
Our fusion strategy further boosts accuracy on out-of-distribution classes, as demonstrated in \cref{tab:exp:potsdamclass}. Most CLIP-based methods and their training-free variants struggle with low-resolution data. {SegEarth-OV} and PC attempt to mitigate this with upsamplers and VFM features to improve spatial awareness. However, integrating LandSegmenter’s encoder in PC yields even greater gains, demonstrating the superior feature extraction capabilities of our model.

\begin{figure*}   \includegraphics[width=1.\linewidth]{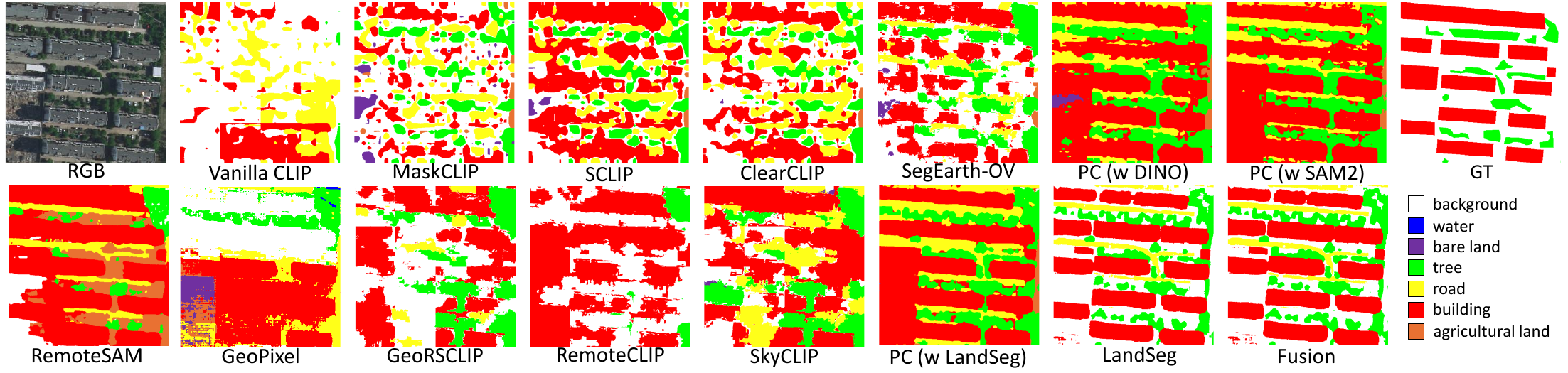}
    \caption{\cy{Segmentation maps generated by various methods on the LoveDA dataset, where LandSeg denotes LandSegmenter.}}
    \label{fig:cmap:loveda}
\end{figure*}

\begin{figure*}
\includegraphics[width=1.\linewidth]{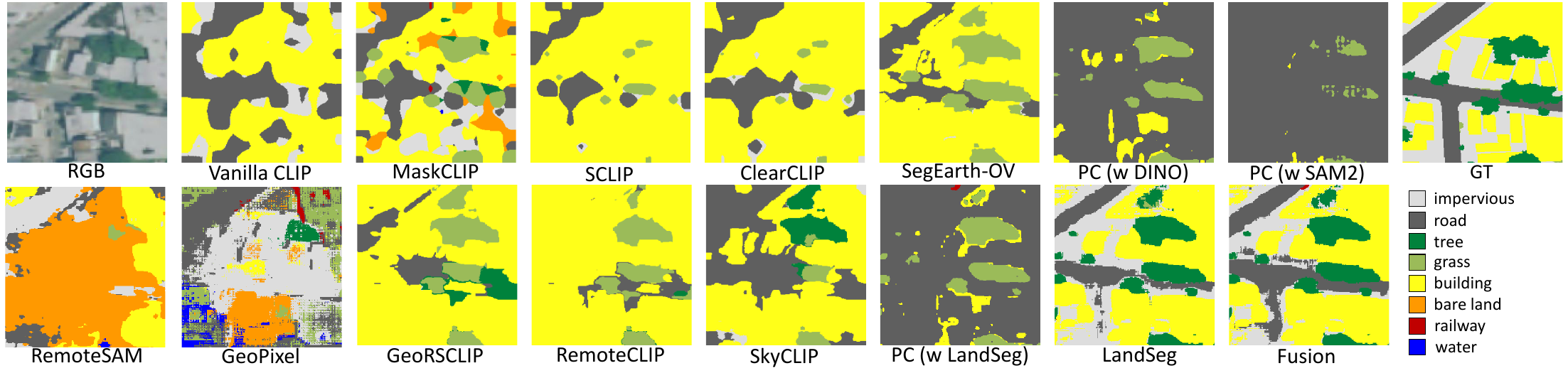}
    \caption{\cy{Segmentation maps generated by various methods on the NYC dataset, where LandSeg denotes LandSegmenter.}}
    \label{fig:cmap:nyc}
\end{figure*}

\begin{figure*}
\includegraphics[width=1.\linewidth]{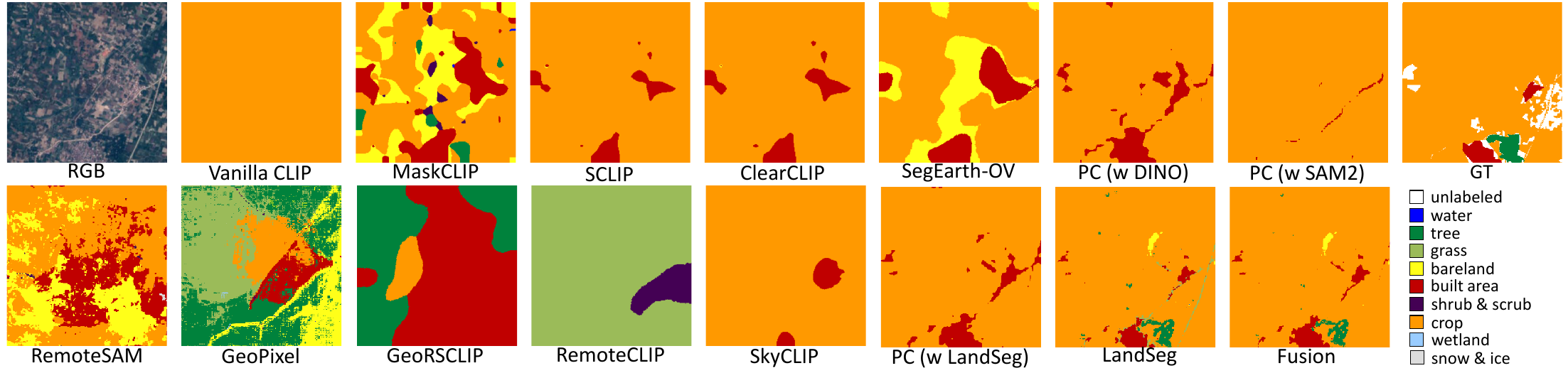}
    \caption{\cy{Segmentation maps generated by various methods on the DW dataset, where LandSeg denotes LandSegmenter.}}
    \label{fig:cmap:dw}
\end{figure*}

\begin{figure*}
    \includegraphics[width=1.\linewidth]{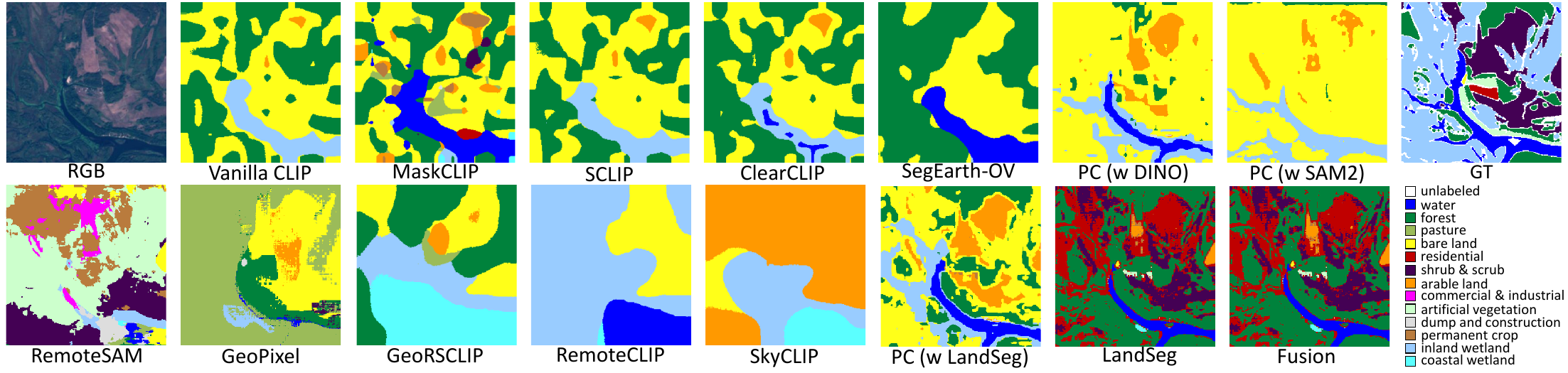}
    \caption{\cy{Segmentation maps generated by various methods on the OSM dataset, where LandSeg denotes LandSegmenter.}}
    \label{fig:cmap:osm}
\end{figure*}

\begin{figure*}
    \includegraphics[width=1.\linewidth]{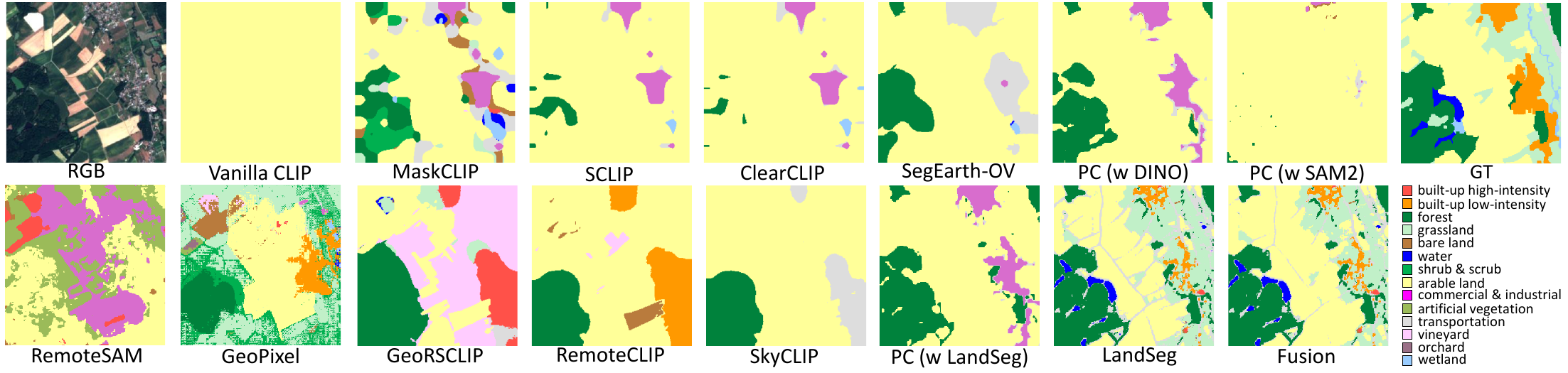}
    \caption{\cy{Segmentation maps generated by various methods on the MultiSenGe dataset, where LandSeg denotes LandSegmenter.}}
    \label{fig:cmap:multisenge}
\end{figure*}

For qualitative assessment, we present example segmentation maps produced by various methods in \cref{fig:cmap:loveda}–\cref{fig:cmap:multisenge}. These visual comparisons highlight the strengths of LandSegmenter, whose LULC maps consistently preserve finer details and exhibit more accurate semantics. The advantages are particularly pronounced on medium-to-low-resolution datasets. 
{In contrast, vanilla CLIP, trained at the image level, captures only coarse semantics with significant spatial detail loss. Other CLIP-based models, especially those incorporating VFM features, partially improve spatial consistency on high-resolution datasets. The three RS-specific CLIP variants, including RemoteCLIP, GeoRSCLIP, and SkyCLIP, also show limited gains, but still underperform the proposed LandSegmenter. RemoteSAM successfully identifies most building areas in~\cref{fig:cmap:loveda}, while GeoPixel only detects buildings from the lower part and trees from the upper part. As GeoPixel is designed for referring RS image segmentation, it struggles to segment all objects without detailed location input. All compared models generalize poorly to low-resolution datasets due to the lack of cross-resolution training data. These results further indicate the effectiveness of the proposed method.}

\subsubsection{Fine-tuning}

We assess LandSegmenter’s fine-tuning performance under minimal supervision in \cref{tab:exp:ft}. In this setting, we compare LandSegmenter with six state-of-the-art RS VFMs, each designed for different input band configurations.
Besides, we also consider three SAM2 variants for comparison. LandSegmenter achieves competitive or superior results across all datasets. LandSegmenter’s flexibility enables it to process diverse datasets without band limitations and the need to change the classification header. Other compared methods have more or less restricted applicability. Comparisons of SAM2+HR and SAM2+HR+DOFA* reveal the advantage of spectral integration for multispectral data. Their instability shows LAS’s essential role in pretraining for robust fine-tuning.

\begin{table*}[ht]
\centering
\footnotesize
\caption{Fine-tuning results on six test datasets, where DOFA* refers to the original DOFA weights, while the DOFA models in LandSegmenter are further fine-tuned on the LAS dataset. We utilize UperNet~\citep{xiao_unified_2018} and DeepLabv3+~\citep{chen_encoder-decoder_2018} as the segmentation frameworks for ViT-Large (scaleMAE, satMAE, satMAE++, DOFA) and ResNet50 (CromSS, DeCUR) backbones. In SAM2-related models, we fix the backbones (SAM2 and DOFA) during the fine-tuning. For compared methods, all the weights are adjusted. \textsuperscript{†} Failure fine-tuning cases.}
\setlength{\extrarowheight}{0.mm}
\setlength\tabcolsep{3.2pt}
\begin{tabular}{c||c|c||c|c||c|c||c|c||c|c||c|c}
\toprule
\textbf{Dataset}        & \multicolumn{2}{c||}{\textbf{Potsdam}} & \multicolumn{2}{c||}{\textbf{LoveDA}} & \multicolumn{2}{c||}{\textbf{NYC}} & \multicolumn{2}{c||}{\textbf{DW}} & \multicolumn{2}{c||}{\textbf{OSM}} & \multicolumn{2}{c}{\textbf{MultiSenGe}} \\
\midrule
\textbf{Training size}     & 0.1               & 0.3              & 0.1              & 0.3              & 0.1             & 0.3            & 0.1            & 0.3            & 0.1             & 0.3            & 0.1                & 0.3                \\
\midrule
scaleMAE (3B)~\citep{reed_scale-mae_2023}    & 63.29          & 67.04          & {\ul 49.46}    & {\ul 50.96}    & \textbf{56.96} & \textbf{61.88} & 52.23                                    & 55.21          & 32.20                                    & 37.01          & 36.75          & 39.82          \\
satMAE (3B)~\citep{cong_satmae_2022}         & 60.31          & 64.48          & 47.08          & 48.85          & {\ul 55.81}    & {\ul 61.01}    & 52.90                                    & 56.87          & 34.44                                    & 37.76          & 34.17          & 37.76          \\
satMAE (10B)~\citep{cong_satmae_2022}        & -              & -              & -              & -              & -              & -              & 46.61                                    & 51.44          & 29.91                                    & 33.60          & 32.35          & 36.15          \\
satMAE++ (3B)~\citep{noman_rethinking_2024}  & 58.47          & 62.91          & 44.38          & 45.77          & 54.18          & 59.57          & 39.70                                    & 52.89          & 22.89                                    & 34.44          & 32.08          & 34.17          \\
satMAE++ (10B)~\citep{noman_rethinking_2024} & -              & -              & -              & -              & -              & -              & 50.16                                    & 54.72          & 32.26                                    & 37.45          & 33.16          & 36.17          \\
CromSS (9B)~\citep{liu_cromss_2025}          & -              & -              & -              & -              & -              & -              & 57.89                                    & 58.08          & 36.31                                    & 37.33          & 29.04          & 36.37          \\
CromSS (13B)~\citep{liu_cromss_2025}         & -              & -              & -              & -              & -              & -              & {\ul 59.33}                              & 59.38          & 36.34                                    & {\ul 42.47}    & -              & -              \\
DeCUR (13B)~\citep{wang_decur_2023}          & -              & -              & -              & -              & -              & -              & 54.81                                    & 55.78          & {\ul 37.50}                              & 41.57          & -              & -              \\
DOFA~\citep{xiong_neural_2024}               & 61.85          & 65.97          & 47.03          & 48.62          & 54.29          & 60.22          & 54.99                                    & 55.66          & 35.46                                    & 39.11          & {\ul 36.76}    & 40.10          \\ \hline
SAM2 (3B)                                                            & 52.27          & 59.82          & 42.99          & 45.24          & 32.46          & 43.49          & 45.82                                    & 52.61          & 24.54                                    & 28.80          & 24.36          & 29.02          \\
SAM2+HR (3B)                                                         & {\ul 67.09}    & {\ul 71.01}    & 47.57          & 49.83          & 44.81          & 53.64          & 7.88\textsuperscript{†} & 60.24          & 26.07                                    & 35.09          & 33.08          & 40.58          \\
SAM2+HR+DOFA*                                                        & 66.59          & 70.88          & 47.00          & 50.16          & 45.48          & 57.45          & 4.07\textsuperscript{†} & \textbf{61.73} & 1.62\textsuperscript{†} & 35.78          & 35.41          & {\ul 42.12}    \\ \hline
\textbf{LandSegmenter}                                               & \textbf{69.16} & \textbf{71.56} & \textbf{50.74} & \textbf{51.77} & 54.48          & 59.41          & \textbf{60.33}                           & {\ul 60.88}    & \textbf{43.46}                           & \textbf{44.80} & \textbf{41.40} & \textbf{44.75}\\
\bottomrule
\end{tabular}
\label{tab:exp:ft}
\end{table*}

\subsection{Ablation study}

We conduct ablation experiments to examine the impact of weak supervision, architectural components, and training strategies in LandSegmenter construction, with a particular focus on their contributions to zero-shot performance.

\subsubsection{Role of weak labels}

\begin{table*}[t]
\footnotesize
\caption{Zero-shot segmentation performance (mIoU) by LandSegmenter trained with different data partitions, where \textbf{W}, \textbf{E}, \textbf{S}, and \textbf{L} represent the six weak subsets, two exact subsets, three S2 subsets, and three Landsat subsets, respectively.}
\setlength{\extrarowheight}{0.mm}
\setlength\tabcolsep{8pt}
\begin{tabular}{c|cccc>{\columncolor[HTML]{E2E2E2}}c|cccc>{\columncolor[HTML]{E2E2E2}}c}
\toprule
\textbf{Method}             & \multicolumn{5}{c|}{\textbf{LandSegmenter}}              & \multicolumn{5}{c}{\textbf{Confidence-guided Fusion}}               \\
\toprule
\textbf{Training data}             & \textbf{w/o W} & \textbf{w/o E} & \textbf{w/o S} &\textbf{ w/o L} & \textbf{Full set} & \textbf{w/o W} & \textbf{w/o E} & \textbf{w/o S} & \textbf{w/o L} & \textbf{Full set} \\
\midrule
Potsdam    & 39.57 & 7.08  & 41.33 & \textbf{41.65} & {\ul 41.53}    & 48.43 & 21.65 & 47.30 & \textbf{50.66} & {\ul 49.73}    \\
LoveDA     & \textbf{41.32} & 3.43  & 39.77 & 40.09 & {\ul 40.40}    & \textbf{41.40} & 8.76  & 40.21 & 40.49 & {\ul 40.87}    \\
NYC        & 25.40 & 4.38  & 30.74 & \textbf{31.62} & {\ul 31.44}    & 32.46 & 9.28  & {\ul 33.55} & \textbf{35.31} & 33.34    \\
DW         & 18.05 & \textbf{44.39} & 32.54 & 41.36 & {\ul 44.08}    & 25.31 & \textbf{46.71} & 36.96 & 42.85 & {\ul 46.06}    \\
OSM        & 11.05 & {\ul 28.95} & 19.45 & 26.30 & \textbf{29.35}    & 15.26 & 30.14 & 22.54 & {\ul 27.99} & \textbf{30.69}    \\
MultiSenGe & 9.13  & 15.36 & 9.79  & {\ul 15.94} & \textbf{18.07}    & 12.26 & 16.34 & 11.16 & {\ul 17.18} & \textbf{18.92}    \\
\midrule
\textbf{Average}    & 24.09 & 17.27 & 28.94 & {\ul 32.83} & \textbf{34.15}    & 29.19 & 22.15 & 31.95 & {\ul 35.75} & \textbf{36.60}   \\
\bottomrule
\end{tabular}
\label{tab:exp:abla:data}
\end{table*}

We evaluate the role of weak labels by selectively excluding different data partitions of LAS during LandSegmenter training. As shown in \cref{tab:exp:abla:data}, using the full LAS dataset yields the best balanced performance across all benchmarks. In LAS, exact (E) and weak (W) label sets correspond to high- and low-resolution imagery, respectively. Excluding either subset degrades performance for the associated resolution, indicating the importance of multi-modal input during training. 
{Notably, removing W leads to substantial performance drops on S2 test sets including DW, OSM, and MultiSenGe, demonstrating the value of weak labels in pretraining. Similarly, excluding the S2-aligned subsets (S) also significantly reduces accuracy on S2 test sets, although the drop is less severe than when both S and Landsat-aligned (L) subsets are excluded. These findings demonstrate the effectiveness and robustness of using weak labels from LULC products for FM training.} 
Interestingly, omitting S has a larger impact on Potsdam, LoveDA, and NYC than excluding L, suggesting stronger interactions and transferability among datasets with similar resolutions. Another key observation is that the confidence-guided fusion strategy helps mitigate performance gaps caused by partial training data. This finding highlights the effectiveness of the proposed fusion mechanism in enhancing model’s robustness and generalization.

\begin{figure}
    \centering
    \includegraphics[width=1\linewidth]{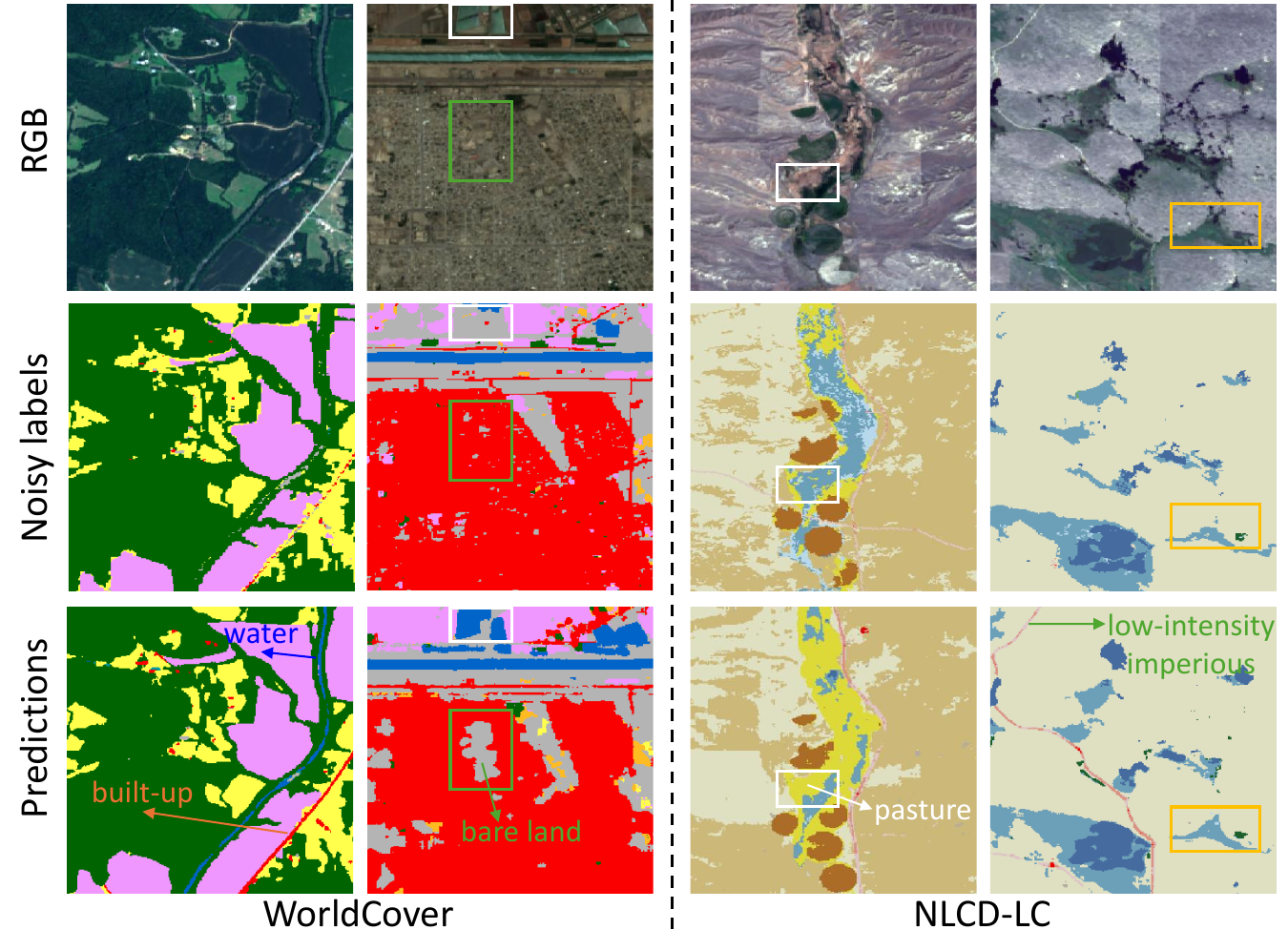}
    \caption{{Visual comparison between noisy labels and LandSegmenter predictions from the WorldCover and NLCD-LC training sets. The detailed color scheme is available in the supplementary material.}}
    \label{exp:fig:trainexample}
\end{figure}

{We further provide visual comparisons between LandSegmenter's predictions and the noisy training labels in \cref{exp:fig:trainexample}. Despite the label noise in the training data, LandSegmenter can avoid overfitting to mislabeled regions and preserve fine spatial details. For example, missing rivers and roads are delineated in the prediction masks. Some mismatches caused by seasonal changes (e.g., wetland to pasture) are also corrected. These results reinforce the effectiveness of these weak labels in large-scale FM training.}

\subsubsection{Architecture design and training strategies} \label{sec:exp:ablation:arch}

\begin{table*}[htp]
\footnotesize
\caption{Zero-shot segmentation performance (mIoU) of the models with different components and training strategies. \textsuperscript{†} Learning rate scale for fine-tuning DOFA.}
\setlength{\extrarowheight}{0.2mm}
\setlength\tabcolsep{4.1pt}
\begin{tabular}{c|cccccc>{\columncolor[HTML]{E2E2E2}}c|cccccc>{\columncolor[HTML]{E2E2E2}}c}
\toprule
\textbf{Method}     & \multicolumn{7}{c|}{\textbf{LandSegmenter}}                                                                                                                                                                                   & \multicolumn{7}{c}{\textbf{Confidence-guided Fusion}} \\
\midrule
\textbf{HR extractor}         & \textbf{\ding{55}} & \textbf{\ding{51}} & \textbf{\ding{55}} & \textbf{\ding{51}}                & \textbf{\ding{51}}                & \textbf{\ding{51}}                  & \textbf{\ding{51}} &\textbf{\ding{55}} & \textbf{\ding51} & \textbf{\ding{55}} & \textbf{\ding{51}}                & \textbf{\ding{51}}                & \textbf{\ding{51}}  &     \textbf{\ding{51}}   \\
\textbf{DOFA}       & \textbf{\ding{55}} & \textbf{\ding{55}} & \textbf{0.1\textsuperscript{†}} & \textbf{0\textsuperscript{†}} & \textbf{1\textsuperscript{†}} & \textbf{0.1\textsuperscript{†}} & \textbf{0.1\textsuperscript{†}} & \textbf{\ding{55}} & \textbf{\ding{55}} & \textbf{0.1\textsuperscript{†}} & \textbf{0\textsuperscript{†}} & \textbf{1\textsuperscript{†}} & \textbf{0.1\textsuperscript{†}} &  \textbf{0.1\textsuperscript{†}} \\
\textbf{Auxiliary decoder}    &  \textbf{\ding{51}} & \textbf{\ding{51}} & \textbf{\ding{51}} & \textbf{\ding{51}} & \textbf{\ding{51}} & \textbf{\ding{55}} & \textbf{\ding{51}} &   \textbf{\ding{51}} & \textbf{\ding{51}} & \textbf{\ding{51}} & \textbf{\ding{51}} & \textbf{\ding{51}} & \textbf{\ding{55}} & \textbf{\ding{51}} \\ 
\midrule
Potsdam    & 26.34                  & 35.05                  & {\ul 41.13}                           & 37.50                                  & 36.97 & 38.30                                    & \textbf{41.53}         & 40.76                  & 46.99                                 & 48.37                                 & 48.51          & \textbf{49.91}                          & 48.26          & {\ul 49.73}    \\
LoveDA     & 35.09                  & 40.31                  & \textbf{43.11}                        & {\ul 42.52}                           & 39.46 & 39.47                                   & 40.55                  & 37.03                  & 40.37                                 & \textbf{43.44}                        & {\ul 43.08}    & 39.61                                   & 39.25          & 40.81          \\
NYC        & 19.53                  & 15.46                  & 26.76                                 & 22.80                                  & 14.07 & {\ul 28.76}                             & \textbf{31.44}         & 21.67                  & 18.65                                 & 31.41                                 & 25.17          & 18.07                                   & \textbf{33.78} & {\ul 33.34}    \\
DW         & 39.26                  & 43.86                  & 41.50                                 & \textbf{46.65}                        & 41.11 & 43.77                                   & {\ul 44.08}            & 19.29                  & 36.07                                 & 43.78                                 & \textbf{49.59} & 44.03                                   & 45.40           & {\ul 46.06}    \\
OSM        & 23.11                  & 25.31                  & 25.18                                 & 27.92                                 & 24.54 & {\ul 28.99}                             & \textbf{29.35}         & 23.83                  & 16.40                                  & 26.44                                 & 29.15          & 25.79                                   & {\ul 29.81}    & \textbf{30.69} \\
MultiSenGe & 11.61                  & 15.38                  & 15.57                                 & 15.99                                 & 16.49 & \textbf{18.65}                          & {\ul 18.07}            & 12.18                  & 13.22                                 & 17.52                                 & 17.19          & 18.53                                   & \textbf{19.59} & {\ul 18.92}    \\
\midrule
Average    & 25.82                  & 29.23                  & 32.21                                 & 32.23                                 & 28.77 & {\ul 32.99}                             & \textbf{34.17}         & 25.79                  & 28.62                                 & 35.16                                 & 35.45          & 32.66                                   & {\ul 36.02}    & \textbf{36.59} \\
\bottomrule
\end{tabular}
\label{tab:exp:abla:arch}
\end{table*}

We evaluate the contributions of LandSegmenter’s architectural components in \cref{tab:exp:abla:arch}. The full model, combined with our tailored training strategy, achieves the best overall performance. Incorporating the adapter with HR extractors significantly improves results, highlighting both the domain gap of SAM2 on RS imagery and the importance of spatial detail in segmentation tasks. Integrating DOFA to leverage spectral information further boosts accuracy. However, aggressive fine-tuning with a high learning rate can lead to overfitting and reduced generalization. Using fixed DOFA weights without fine-tuning during LandSegmenter training still yields strong performance, especially when combined with the proposed confidence-guided fusion strategy. This indicates the effectiveness of existing FMs and the benefit of integrating multiple FMs tailored to specific downstream tasks. Moreover, introducing an auxiliary decoder during training leads to consistent gains across the datasets, demonstrating the effectiveness of this simple training technique. Overall, these findings validate the design choices in both the model architecture and training pipeline of LandSegmenter.


\begin{figure}
    \includegraphics[width=1.\linewidth]{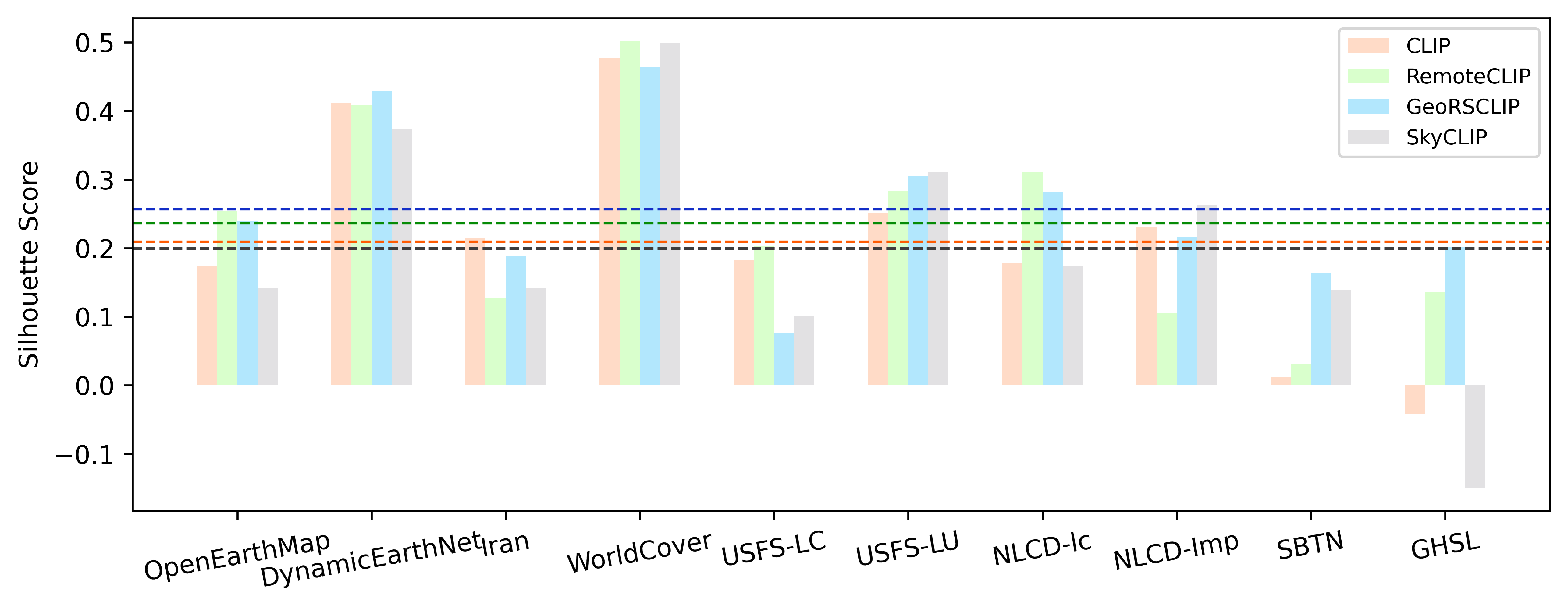}
    \caption{Silhouette scores of text embeddings generated by the text encoders of CLIP, RemoteCLIP, GeoRSCLIP, and SkyCLIP. Dashed lines indicate the mean score values across the datasets. Higher is better.}
\label{fig:exp:sils}
\end{figure}

Then, we evaluate the encoding capability of various CLIP text encoders using Silhouette scores~\citep{rousseeuw_silhouettes_1987}, which measure how well data points (in our case, text embeddings) are clustered based on their semantic similarity. Ranging from -1 to 1, a higher Silhouette score indicates that text embeddings are more tightly grouped within their respective class and well-separated from others, and vice versa. 
Specifically, we generate text embeddings for each training set using all augmented text prompts (see Appendix for the full list). We apply t-SNE for dimensionality reduction prior to computing the Silhouette scores. As shown in \cref{fig:exp:sils}, the text encoder of GeoRSCLIP achieves the highest scores in most cases, followed by RemoteCLIP, which also demonstrates a strong ability to encode LULC knowledge. SkyCLIP’s text encoder struggles to effectively differentiate among complex LULC classes on some datasets. These findings support our choice of GeoRSCLIP’s text encoder. 

\subsubsection{{Region-Level Segmentation Analysis}} \label{sec:exp:region}

\begin{figure}
\centering
    \includegraphics[width=1.\linewidth]{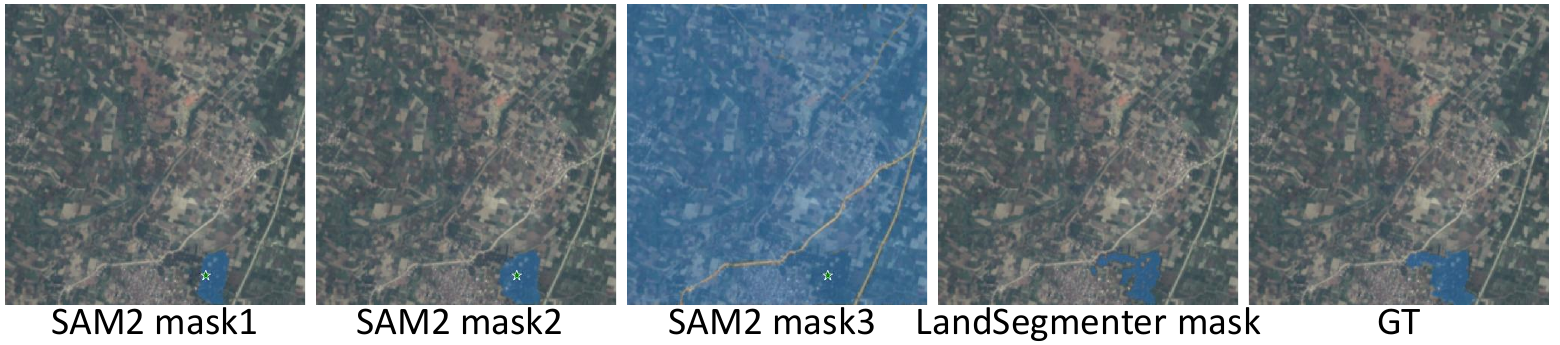}
    \caption{{Comparison of forest segmentation by SAM2 (guided by a point prompt, indicated by stars, producing three candidate masks per query) and LandSegmenter (guided by the class name string). Example from the DW dataset.}}
    \label{exp:fig:compsam}
\end{figure}

{We evaluate LandSegmenter’s performance on region-level classes (e.g., forest, grass, crop), which dominate LULC mapping but lack clear boundaries. As shown in \cref{exp:fig:compsam}, LandSegmenter delineates the forest region more accurately than SAM2. This improvement is also reflected in the class-wise results. On the high-resolution Potsdam dataset (see \cref{tab:exp:potsdamclass}), LandSegmenter achieves higher accuracy on region-level categories such as low vegetation and impervious surfaces. In terms of PC models, replacing SAM2 encoder features with those from LandSegmenter also improves region-level performance.
On the low-resolution DW dataset (\cref{exp:tab:dwcls}), where all categories are region-level, the same tendency holds. These results confirm that LandSegmenter enhances regional consistency and segmentation quality in LULC mapping.}

\begin{table*}[htp]
\centering
\footnotesize
\caption{{Class-wise  zero-shot results on the DW dataset.}}
\setlength{\extrarowheight}{0.mm}
\setlength\tabcolsep{6.8pt}
\begin{tabular}{c|ccccccccc|c}
\toprule
\multirow{2}{*}{\textbf{DW}} & \multicolumn{9}{c|}{\textbf{IoU (\%)}}                                                                          & \multirow{2}{*}{\textbf{mIoU}} \\
\cline{2-10}
                    & \textbf{water} & \textbf{forest} & \textbf{grass} & \textbf{wetland} & \textbf{crop}  & \textbf{shrub} & \textbf{built-up} & \textbf{bare land} & \textbf{ice \& snow} &    \\
\midrule
SegEarth-OV         & 59.12 & 42.41  & 2.79  & 7.98    & 46.23 & 0.62            & 56.46         & 6.56      & 31.94        & 28.23                              \\
PC (w DINO)         & 69.57 & 55.36  & 1.78  & 16.10    & {\ul 65.94} & 0.77     & {\ul 74.82}         & 15.27     & {\ul 37.88}        & 37.50  \\
PC (w SAM2)         & 68.77 & 52.38  & 0.57  & 16.39   & 63.74 & 0.42            & 68.65         & 15.37     & 35.50         & 35.76    \\
\hline
RemoteCLIP          & 56.94 & 23.42  & 2.59  & 4.02    & 40.16 & 13.26           & 41.66         & 10.15     & 23.37        & 23.95   \\
GeoRSCLIP           & 57.29 & 39.82  & 5.69  & 8.50    & 44.24 & 3.44            & 41.41         & 11.57     & 36.28        & 27.58    \\
SkyCLIP             & 43.58 & 27.41  & 0.20  & 6.85    & 35.06 & 0.00               & 54.65         & 10.40      & 37.48        & 23.96    \\
\midrule
PC (w LandSegmenter)      & 72.08 & 61.16  & 2.11  & 22.24   & \textbf{68.65} & 0.37            & \textbf{78.07}         & 16.26     & \textbf{39.07}        & 40.00    \\
LandSegmenter             & {\ul 82.83} & {\ul 73.98}  & {\ul 10.28} & {\ul 34.26}   & 57.54 & {\ul 16.23}           & 66.77         & {\ul 44.40}      & 10.45        & {\ul 44.08}   \\
Fusion              & \textbf{83.42} & \textbf{74.58}  & \textbf{10.97} & \textbf{35.30}    & 61.80  & \textbf{16.37}           & 71.90          & \textbf{45.00}       & 15.26        & \textbf{46.06}   \\
\bottomrule
\end{tabular}
\label{exp:tab:dwcls}
\end{table*}

\subsection{Hyperparameter sensitivity} \label{sec:exp:hyper}

Finally, we investigate the impact of the confidence threshold $C_t$ on the effectiveness of the fusion strategy. As shown in \cref{exp:tab:ct}, the proposed confidence-guided fusion method is robust to variations in this hyperparameter. Nevertheless, a slightly higher threshold above 0.5 tends to yield better performance.

\begin{table}[htp]
\footnotesize
\caption{Zero-shot results (mIoU) using the confidence-guided fusion with varying $C_t$, where we use abbreviations for Potsdam (Pd), LoveDA (LD), MultiSenGe (MSG), and Average (Avg.).}
\setlength{\extrarowheight}{0.mm}
\setlength\tabcolsep{4.8pt}
\begin{tabular}{c|cccccc|c}
\toprule
\textbf{$C_t$} & \textbf{Pd}    & \textbf{LD}    & \textbf{NYC}   & \textbf{DW}    & \textbf{OSM}   & \textbf{MSG}   & \textbf{Avg.}  \\
\midrule
0.4   & 46.61 & 41.00 & 33.41 & 46.23 & 30.62 & 19.01 & 36.15 \\
0.5   & 47.21 & 40.93 & 33.37 & 46.26 & 30.57 & 19.00 & 36.22 \\
0.6   & 49.73 & 40.87 & 33.34 & 46.06 & 30.69 & 18.92 & 36.60 \\
0.7   & 49.84 & 40.83 & 33.28 & 45.94 & 30.24 & 18.94 & 36.51 \\
0.8   & 49.62 & 40.76 & 33.25 & 45.70 & 30.36 & 18.87 & 36.43 \\
\bottomrule
\end{tabular}
\label{exp:tab:ct}
\end{table}

\section{Conclusions} \label{sec:conclusion}

We propose LandSegmenter, an LULC FM of high input-output flexibility. For its training, we curate the LAS dataset, a large-scale, multi-modal collection predominantly weakly labeled by LULC products, offering a cost-efficient alternative to expensive manual annotations. We also introduce the confidence-guided fusion strategy to boost zero-shot inference. Transfer learning experiments across six diverse LULC datasets demonstrate LandSegmenter’s effectiveness, particularly on low-resolution multispectral imagery and zero-shot settings, showing the potential of weak labels in scaling up FM construction.
However, this work represents our first step toward building a unified, task-specific LULC FM. Several challenges remain and require further investigation. One notable challenge is the development of more effective strategies to mitigate label noise. Directly integrating advanced learning from noisy labels (LNL) methods into LULC FM training is non-trivial, as most approaches rely on multi-round~\citep{liu_early-learning_2020}, multi-model~\citep{han_co-teaching_2018}, or multi-input~\citep{li_dividemix_2020} learning strategies. Implementing these strategies with pixel-level FMs would significantly increase storage and computational demands due to repeated storage of intermediate results, concurrent training of multiple encoder-decoder models, or processing multiple inputs per iteration. Therefore, lightweight noise-robust strategies are required to enable efficient integration into FM training. 
Another key challenge is the inherent class imbalance, amplified by the hierarchical nature of LULC classification. While class-wise reweighting according to sample sizes can potentially mitigate this issue, LandSegmenter also need to balance hierarchical semantics and integrate multimodal visual–textual information, making simple data-level adjustments insufficient. 
Additionally, the current framework employs a fixed text encoder due to the limited text corpus in LAS. \cy{As a result, it is less suited to resolving finer-grained ambiguity within a prompt set, such as synonymy or hierarchical overlap among queried classes. Future work will explore ambiguity-aware text modeling and fine-tuning strategies to improve the understanding of hierarchical LULC semantics. This will require both specialized data collection and tailored training approaches.}

\section*{Acknowledgement}
This project is jointly supported by the Munich Center for Machine Learning and the German Research Foundation (DFG GZ: ZH 498/18-1; Project number: 519016653).

\section*{Author Contributions}
Chenying Liu: Conceptualization, Methodology, Experiments, Software, Validation, Formal analysis, Investigation, Data Curation, Visualization, Writing - Original Draft, Review \& Editing; Wei Huang: Methodology, Experiments, Formal analysis, Writing - Review \& Editing; Xiao Xiang Zhu: Conceptualization, Methodology, Writing - Review \& Editing, Project administration, Supervision, Funding acquisition.

\section*{Declaration of generative AI and AI-assisted technologies in the writing process}

During the preparation of this work, the authors used ChatGPT in order to improve readability and language. After using this tool/service, the authors reviewed and edited the content as needed and take full responsibility for the content of the publication.

\bibliographystyle{cas-model2-names}
\bibliography{main}


\def\tsc#1{\csdef{#1}{\textsc{\lowercase{#1}}\xspace}}
\tsc{WGM}
\tsc{QE}





\appendix
\clearpage 
\onecolumn

\renewcommand{\thefigure}{A.\arabic{figure}}  
\renewcommand{\thetable}{A.\arabic{table}}    
\setcounter{figure}{0}
\setcounter{equation}{0}
\setcounter{table}{0}

\begin{center}
  {\LARGE \bfseries Appendix of LandSegmenter: Towards a Unified Foundation Model for Land Use and Land
Cover Mapping}
\end{center}





\section{SLA dataset} \label{app:data}

\subsection{Class systems and text prompts} \label{app:data:classnames}

Below, we detail the class systems and the corresponding class name text prompts used for training across the eight LAS subsets, as listed in \cref{app:tab:data:oem,app:tab:data:den,app:tab:data:iran,app:tab:data:ghsl,app:tab:data:worldcover,app:tab:data:nlcd-lc,app:tab:data:nlcd-imp,app:tab:data:usfs-lc,app:tab:data:usfs-lu,app:tab:data:sbtn}. With NLCD and USFS each having two layer types, this results in 10 class systems. Each table includes original class names from the LULC products and our reorganized text strings. To standardize class names across subsets, we mainly follow three rules:
\begin{itemize}
\item using uniform descriptions for identical definitions (e.g., `water', `open water', and `water body' are unified as `water' and `lakes, reservoirs, rivers, and oceans');
\item trying to keep consistent granularity across subsets, using connectors like `and', `except for', and `including' to describe mixed classes (e.g., using `except for' to define the relationship between `developed area' and `building': \textit{developed area except for building}), hoping the model able to learn the simple connections to some extent;
\item changing all plural forms to singular for consistency purposes.
\end{itemize}

\newpage

\begin{table*}
\centering
\small
\caption{Class information and corresponding used name texts for the OpenEarthMap~\citep{xia_openearthmap_2023} subset.}
\setlength{\extrarowheight}{0.3mm}
\setlength\tabcolsep{3.2pt}
\begin{tabular}{cccp{2.0cm}p{9.8cm}}
\toprule
\textbf{Class} & \textbf{Color}    & \textbf{Proportion} &\textbf{Original name}    & \textbf{Used name text}                                                                                                                                                                                                                                 \\
\midrule
0     & \cellcolor[HTML]{800000}\textcolor{white}{\#800000} & 1.51       & Bareland         & barren land; bare   land; rock, sand, clay and soil                                                                                                                                                                                             \\
\hline
1     & \cellcolor[HTML]{00FF24}\#00FF24 & 22.67      & Rangeland        & rangeland and herbaceous wetland; grass,   pasture, scrub and herbaceous wetland; herbaceous vegetation, shrub and   herbaceous wetland; grass, pasture and shrub and herbaceous wetland;   herbaceous vegetation, scrub and herbaceous wetland \\
\hline
2     & \cellcolor[HTML]{949494}\#949494 & 15.37      & Developed space  & developed area except for building;   built-up area except for building                                                                                                                                                                         \\
\hline
3     & \cellcolor[HTML]{7030A0}\textcolor{white}{\#7030A0} & 6.62       & Road             & road; transportation                                                                                                                                                                                                                            \\
\bottomrule
\end{tabular}
\label{app:tab:data:oem}
\end{table*}

\begin{table*}
\centering
\small
\caption{Class information and corresponding used name texts for the DynamicEarthNet~\citep{toker_dynamicearthnet_2022} subset.}
\setlength{\extrarowheight}{0.3mm}
\setlength\tabcolsep{3.2pt}
\begin{tabular}{cccp{2.4cm}p{9.6cm}}
\toprule
Class & Color    & Proportion & Original name               & Used name text    \\
\midrule
0     & \cellcolor[HTML]{949494}\#949494 & 7.16       & impervious surface          & developed impervious area; built-up   impervious area                                                 \\
\hline
1     & \cellcolor[HTML]{CCCC00}\#CCCC00 & 9.55       & agriculture                 & agricultural land; crop; crop land                                                                    \\
\hline
2     & \cellcolor[HTML]{00FF24}\#00FF24 & 45.49      & forest and other vegetation & forest, grass, shrub, pasture and   artificial vegetation; tree, herbaceous vegetation and scrub and artificial   vegetation; tree, herb, shrub and artificial vegetation; tree, herbaceous   vegetation, shrub and artificial vegetation; tree, grass, shrub, pasture and   artificial vegetation; forest, herbaceous vegetation, scrub and artificial   vegetation; forest, herb, shrub and artificial vegetation \\
\hline
3     & \cellcolor[HTML]{2809C3}\textcolor{white}{\#2809C3} & 0.72       & wetland                     & wetland                                                                                               \\
\hline
4     & \cellcolor[HTML]{B57917}\#B57917 & 28.06      & soil                        & barren land; bare   land; rock, sand, clay and soil                                                   \\
\hline
5     & \cellcolor[HTML]{0594C7}\#0594C7 & 8.04       & water                       & water; lakes,   reservoirs, rivers and ocean                                                           \\
\hline
6     & \#FFFFFF & 0.99       & snow and ice                & snow and ice             \\
\bottomrule
\end{tabular}
\label{app:tab:data:den}
\end{table*}

\begin{table*}[h]
\small
\centering
\caption{Class information and corresponding used name texts for the Iran (\href{https://developers.google.com/earth-engine/datasets/catalog/KNTU_LiDARLab_IranLandCover_V1}{Iran Land Cover Map}) subset.}
\setlength{\extrarowheight}{0.3mm}
\setlength\tabcolsep{3.2pt}
\begin{tabular}{cccp{2cm}p{9.8cm}}
\toprule
\textbf{Class}           & \textbf{Color}                     & \textbf{Proportion}            & \textbf{Original name}   & \textbf{Used name text} \\
\midrule
0                  & \cellcolor[HTML]{000000}\textcolor{white}{\#000000}                  & 4.59                   & Urban           & urban area; developed area;   residential, commercial, industrial and transportation area; built-up area \\
\hline
1                  & \cellcolor[HTML]{006eff}\#006eff                  & 0.58                   & Water           & water; lakes, reservoirs, rivers and ocean\\
\hline
2                  & \cellcolor[HTML]{41a661}\#41a661                  & 6.61                   & Wetland         & wetland except for   marshland  \\
\hline
\multirow{4}{*}{3} & \cellcolor[HTML]{732600} & \multirow{4}{*}{38.52} & Kalut (yardang) & \multirow{4}{*}{\makecell[l]{barren land except for salty land and sand;\\bare land except for salty land and sand;\\rock, clay and soil except for salty land}}     \\
                   & \cellcolor[HTML]{732600}       &                        & Clay            &                                                                                                                                                            \\
                   &  \cellcolor[HTML]{732600}  &                        & Outcrop         &                                                                                                                                                            \\
                   &  \multirow{-4}{*}{\cellcolor[HTML]{732600}\textcolor{white}{\#732600}}  &                        & Uncovered Plain &                                                                                                                                                            \\
\bottomrule
\end{tabular}
\label{app:tab:data:iran}
\end{table*}

\begin{table*}[htp]
\small
\centering
\caption{Class information and corresponding used name texts for the GHSL (\href{https://developers.google.com/earth-engine/datasets/catalog/JRC_GHSL_P2023A_GHS_BUILT_S_10m}{Global Human Settlement Layer - Global built-up surface 10m}) subset. Unlisted portions denote no data and are masked during training.}
\setlength{\extrarowheight}{0.3mm}
\setlength\tabcolsep{3.2pt}
\begin{tabular}{cccp{7.8cm}p{4cm}}
\toprule
\textbf{Class}           & \textbf{Color}                     & \textbf{Proportion}            & \textbf{Original name}                                                                    & \textbf{Used name text}                                                                                                                       \\
\midrule
\multirow{4}{*}{0} & \cellcolor[HTML]{226126} & \multirow{4}{*}{15.34} & open spaces, low vegetation surfaces                                            & \multirow{4}{*}{\makecell[l]{non-developed area;\\non-built-up area;\\pervious area}}                                                            \\
                   &   \cellcolor[HTML]{226126}   &                        & open spaces, medium vegetation surfaces                                         &                                                                                                                                  \\
                   &   \cellcolor[HTML]{226126}  &                        & open spaces, high vegetation surfaces                                           &                                                                                                                                  \\
                   &   \multirow{-4}{*}{\cellcolor[HTML]{226126}\textcolor{white}{\#226126}}   &                        & open spaces, water   surfaces                                                   &                                                                                                                                  \\
\hline
\multirow{5}{*}{1} & \cellcolor[HTML]{FFFF00} & \multirow{5}{*}{13.39} & built spaces, residential, building height \textless{}= 3m                      & \multirow{5}{*}{residential area}                                                                                                \\
\hline
\end{tabular}
\label{app:tab:data:ghsl}
\end{table*}

\begin{table*}[htp]
\small
\centering
\caption{Class information and corresponding used name texts for the WorldCover (\href{https://developers.google.com/earth-engine/datasets/catalog/ESA_WorldCover_v100}{ESA WorldCover 10m v100} and \href{https://developers.google.com/earth-engine/datasets/catalog/ESA_WorldCover_v200}{ESA WorldCover 10m v200}) subset.}
\setlength{\extrarowheight}{0.3mm}
\setlength\tabcolsep{3.2pt}
\begin{tabular}{cccp{3cm}p{9cm}}
\toprule
\textbf{Class} & \textbf{Color}    & \textbf{Proportion} & \textbf{Original name }             & \textbf{Used name text}     \\
\midrule
0     & \cellcolor[HTML]{006400}\textcolor{white}{\#006400} & 21.56      & Tree cover                 & tree inclduing  corresponding artificail vegetation and woody crop but except for mangroves;   forest inclduing corresponding artificail vegetation and woody crop but except for mangroves; wood inclduing corresponding artificail vegetation and   woody crop but except for mangroves                                                                                                                                                           \\
\hline
1     & \cellcolor[HTML]{ffbb22}\#ffbb22 & 6.75       & Shrubland                  & shrub and scrub   inclduing corresponding artificail vegetation and woody crop; shrub   inclduing corresponding artificail vegetation and woody crop; scrub   inclduing corresponding artificail vegetation and woody crop   \\
\hline
2     & \cellcolor[HTML]{ffff4c}\#ffff4c & 20.13      & Grassland                  & grassland and pasture   inclduing corresponding artificail vegetation; herb and pasture inclduing   corresponding artificail vegetation; herbaceous vegetation and pasture   inclduing corresponding artificail vegetation; grassland and meadow   inclduing corresponding artificail vegetation; herb and meadow inclduing   corresponding artificail vegetation; herbaceous vegetation and meadow   inclduing corresponding artificail vegetation \\
\bottomrule
\end{tabular}
\label{app:tab:data:worldcover}
\end{table*}

\begin{table*}[htp]
\small
\centering
\caption{Class information and corresponding used name texts for the NLCD-LC (\href{https://developers.google.com/earth-engine/datasets/catalog/USGS_NLCD_RELEASES_2021_REL_NLCD\#description}{USGS National Land Cover Database}-landcover) subset.}
\setlength{\extrarowheight}{0.3mm}
\setlength\tabcolsep{3.2pt}
\begin{tabular}{cccp{4cm}p{8cm}}
\toprule
\textbf{Class} & \textbf{Color}    & \textbf{Proportion} & \textbf{Original name}                  & \textbf{Used name text}    \\
\midrule
0     & \cellcolor[HTML]{466b9f}\#466b9f & 2.02       & Open water                     & water; lakes,   reservoirs, rivers and ocean                                                                                                                                            \\
\hline
1     & \cellcolor[HTML]{d1def8}\#d1def8 & 0.0038     & Perennial ice/snow             & ice and snow                                                                                                                                                                             \\
\hline
2     & \cellcolor[HTML]{dec5c5}\#dec5c5 & 2.15       & Developed, open space          & artificial vegetation; non-argricultural vegetation; artificial or non-argricultural vegetation                                                                                      \\
\hline
3     & \cellcolor[HTML]{d99282}\#d99282 & 1.1        & Developed, low   intensity     & developed   low-intensity imperious area; built-up low-intensity imperious area                                                                                                          \\
\hline
4     & \cellcolor[HTML]{eb0000}\#eb0000 & 0.54       & Developed, medium   intensity  & developed   medium-intensity imperious area; built-up medium-intensity imperious area                                                                                                   \\
\hline
5     & \cellcolor[HTML]{ab0000}\#ab0000 & 0.16       & Developed high   intensity     & developed   high-intensity imperious area; built-up high-intensity imperious area                                                                                                       \\
\bottomrule
\end{tabular}
\label{app:tab:data:nlcd-lc}
\end{table*}

\begin{table*}[htp]
\small
\centering
\caption{Class information and corresponding used name texts for the NLCD-Imp (\href{https://developers.google.com/earth-engine/datasets/catalog/USGS_NLCD_RELEASES_2021_REL_NLCD\#description}{USGS National Land Cover Database}-impervious) subset.}
\setlength{\extrarowheight}{0.3mm}
\setlength\tabcolsep{3.2pt}
\begin{tabular}{cccp{9cm}p{2.8cm}}
\toprule
\textbf{Class}              & \textbf{Color}                     & \textbf{Proportion}             & \textbf{Original name}        & \textbf{Used name text}  \\
\midrule
 & \cellcolor[HTML]{ffff00} &  & Primary road. Interstates and other major roads. Pixels were derived from the 2018 NavStreets Street Data.                                                                                                                   & \multirow{4}{*}{road; transportation}                                                                                                                                                    \\
                   &  \cellcolor[HTML]{ffff00}    &                        & Secondary road. Non-interstate highways. Pixels were   derived from the 2018 NavStreets Street Data.                                                                                                                           &                                                                                                                                                                                          \\
                   &   \cellcolor[HTML]{ffff00}    &                        & Tertiary road. Any two-lane road. Pixels were derived   from the 2018 NavStreets Street Data.                                                                                                                                  &                                                                                                                                                                                          \\
\bottomrule
\end{tabular}
\label{app:tab:data:nlcd-imp}
\end{table*}

\begin{table*}[htp]
\small
\centering
\caption{Class information and corresponding used name texts for the USFS-LC (\href{https://developers.google.com/earth-engine/datasets/catalog/USFS_GTAC_LCMS_v2023-9\#description}{USDA Forest Service Landscape Change Monitoring System}-Land\_Cover) subset.}
\setlength{\extrarowheight}{0.3mm}
\setlength\tabcolsep{3.2pt}
\begin{tabular}{cccp{4.cm}p{7.8cm}}
\toprule
\textbf{Class} & \textbf{Color}    & \textbf{Proportion} & \textbf{Original name}                   & \textbf{Used name text}                                                                                                                                       \\
\midrule
0     & \cellcolor[HTML]{005e00}\textcolor{white}{\#005e00} & 34.01      & Trees                           & tree; forest; wood;   broadleaf and coniferous forest; deciduous and evergreen forest; broadleaf   and coniferous tree; deciduous and evergreen tree \\
\hline
1     & \cellcolor[HTML]{00cc00}\#00cc00 & 0.27       & Shrubs \& Trees   Mix           & mixed shrub and tree   area; mixed scrub and tree area                                                                                               \\
\hline
2     & \cellcolor[HTML]{b3ff1a}\#b3ff1a & 3.74       & Grass/Forb/Herb \&   Trees Mix  & mixed grass and tree   area; mixed herb and tree area; mixed herbaceous and woody area                                                               \\
\hline
3     & \cellcolor[HTML]{99ff99}\#99ff99 & 0.05       & Barren \& Trees Mix           & mixed barren and tree area; mixed barren   and woody area                                                                                            \\
\hline
4     & \cellcolor[HTML]{e68a00}\#e68a00 & 4.28       & Shrubs                          & shrub and scrub;   shrub; scrub                                                                                                                      \\
\hline
5     & \cellcolor[HTML]{ffad33}\#ffad33 & 12.98      & Grass/Forb/Herb \&   Shrubs Mix & mixed grass and shrub   area; mixed herb and shrub area; mixed grass and scrub area; mixed herb and   scrub area                                     \\
\bottomrule
\end{tabular}
\label{app:tab:data:usfs-lc}
\end{table*}

\begin{table*}[htp]
\small
\centering
\caption{Class information and corresponding used name texts for the USFS-LU (\href{https://developers.google.com/earth-engine/datasets/catalog/USFS_GTAC_LCMS_v2023-9\#description}{USDA Forest Service Landscape Change Monitoring System}-Land\_Use) subset.}
\setlength{\extrarowheight}{0.3mm}
\setlength\tabcolsep{3.2pt}
\begin{tabular}{cccp{2.7cm}p{9.5cm}}
\toprule
\textbf{Class} & \textbf{Color}    & \textbf{Proportion} & \textbf{Original name}        & \textbf{Used name text} \\
\midrule
0     & \cellcolor[HTML]{efff6b}\#efff6b & 16.99      & Agriculture          & agricultural land;   crop; cropland; arable land and permanent crop; herbaceous crop and woody   crop; annual crop, orchard and vineyard                                                                                                                                                   \\
\hline
1     & \cellcolor[HTML]{ff2ff8}\#ff2ff8 & 2.89       & Developed            & developed area; urban   area; residential, commercial, industrial and transportation area including   artificial vegetation; built-up area                                                                                                                                                 \\
\hline
2     & \cellcolor[HTML]{1b9d0c}\textcolor{white}{\#1b9d0c} & 36.99      & Forest               & tree and woody   wetland; forest and woody wetland; wood and woody wetland; broadleaf and   coniferous forest, and woody wetland; deciduous and evergreen forest, and   woody wetland; broadleaf and coniferous tree, and woody wetland; deciduous   and evergreen tree, and woody wetland \\
\hline
3     & \cellcolor[HTML]{97ffff}\#97ffff & 0.74       & Non-Forest Wetland   & herbaceous wetland;   non-forest wetland                                                                                                                                                                                                                                                   \\
\bottomrule
\end{tabular}
\label{app:tab:data:usfs-lu}
\end{table*}

\begin{table*}[htp]
\small
\centering
\caption{Class information and corresponding used name texts for the SBTN (\href{https://developers.google.com/earth-engine/datasets/catalog/WRI_SBTN_naturalLands_v1_2020}{SBTN Natural Lands Map-classification}) subset.}
\setlength{\extrarowheight}{0.3mm}
\setlength\tabcolsep{3.2pt}
\begin{tabular}{cccp{4.2cm}p{7.6cm}}
\toprule
\textbf{Class}               & \textbf{Color}                     & \textbf{Proportion}             & \textbf{Original name}                       & \textbf{Used name text}                                                                                                                                                                                                                                                                             \\
\midrule
0                   & \cellcolor[HTML]{246E24}\textcolor{white}{\#246E24}                  & 18.35                  & natural forests                     & tree; forest; wood;   broadleaf and coniferous forest; deciduous and evergreen forest; broadleaf and coniferous tree; deciduous and evergreen tree                                                                                                                                       \\
\hline
1                   & \cellcolor[HTML]{B9B91E}\#B9B91E                  & 22.14                  & natural short vegetation          & rangeland; grass, pasture and scrub;   herbaceous vegetation and shrub; grass, pasture and shrub; herbaceous   vegetation and scrub                                                                                                                                                        \\
\hline
\multirow{2}{*}{2}  & \cellcolor[HTML]{006eff} & \multirow{2}{*}{2.04}  & natural water                       & \multirow{2}{*}{water; lake, reservoir, river and ocean}                                                                          \\
\bottomrule
\end{tabular}
\label{app:tab:data:sbtn}
\end{table*}

\clearpage

\subsection{Text embedding analysis}

\begin{figure*}[htp]
    \centering
    \includegraphics[width=.9\linewidth]{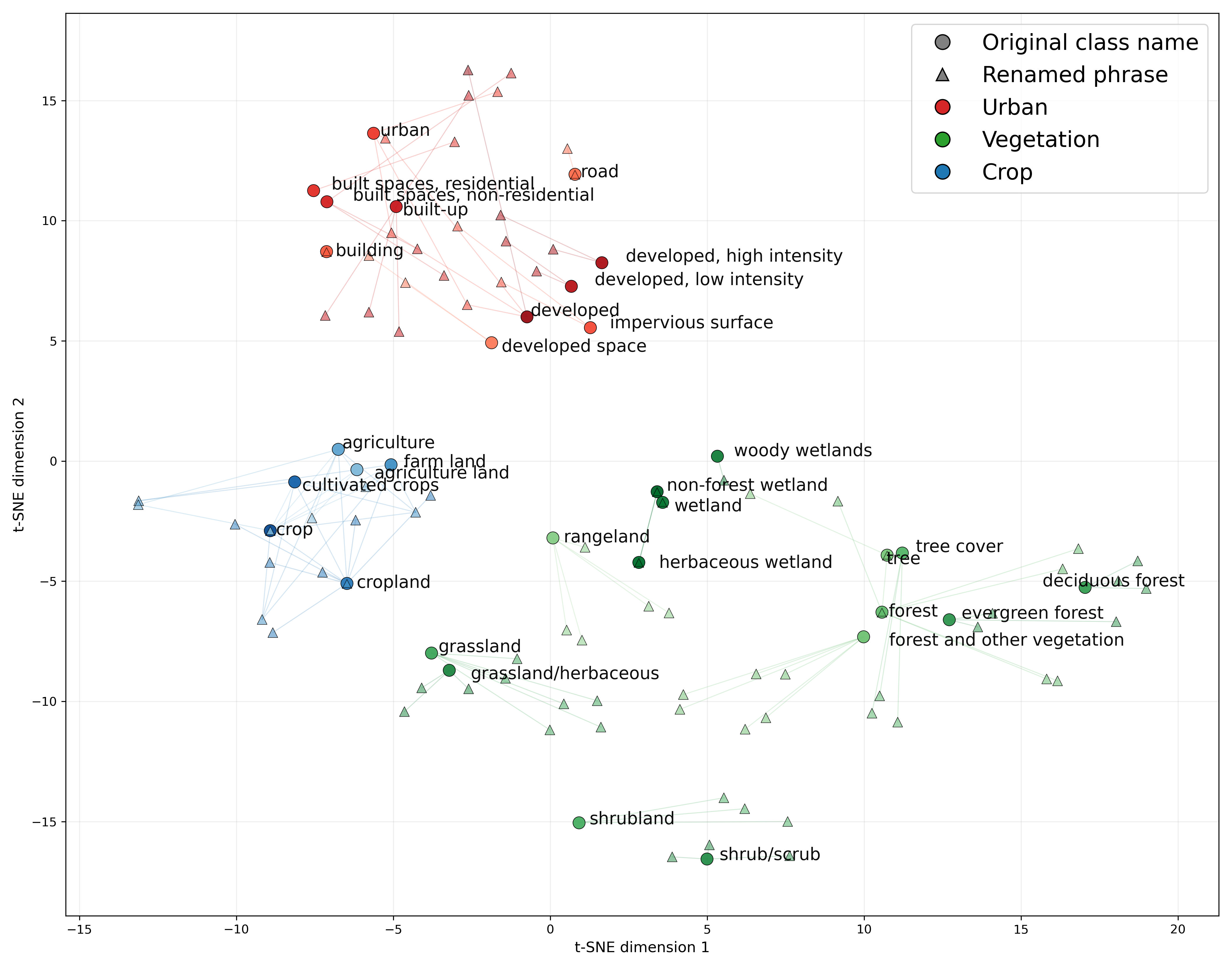}
    \caption{\cy{The t-SNE visualization of GeoRSCLIP text embeddings for original class names and renamed prompts from heterogeneous LAS taxonomies. Semantically related labels from different datasets form coherent clusters in the shared embedding space, while finer-grained concepts remain distinguishable in local subregions.}}
    \label{app:fig:tsne}
\end{figure*}

\cy{The proposed renaming trick acts as a prompt-level semantic alignment strategy. It does not modify the underlying semantic space of the frozen GeoRSCLIP text encoder. Instead, it improves how heterogeneous class labels from different datasets are mapped into and queried within this shared space. To further illustrate this effect, we provide a t-SNE visualization of the text embeddings of both the original class names and their renamed prompts in \cref{app:fig:tsne}. For clarity, we select three representative groups of class-name strings collected from different datasets, namely urban-, crop-, and vegetation-related categories, as listed in \cref{app:tab:data:urban_name}--\cref{app:tab:data:vege_name}. As shown in \cref{app:fig:tsne}, these semantic groups occupy clearly separated regions in the embedding space, indicating good inter-group discrimination. Moreover, within each group, finer-grained subclasses also exhibit meaningful local organization. For example, within the urban-related cluster, terms such as ``high-intensity development'' and ``residential'' are located in nearby yet distinguishable subregions. This is consistent with their semantic roles. The former emphasizes development intensity, while the latter emphasizes land-use function. More general descriptions, such as ``urban'', ``built-up'', and ``developed'' tend to cover broader neighborhoods in the embedding space, especially after prompt expansion through the renaming trick. A similar phenomenon can also be observed in the vegetation-related cluster, where terms such as ``forest'', ``grassland'', ``wetland'', and ``shrubland'' form distinct but semantically related subregions.}

\cy{These observations suggest that the frozen GeoRSCLIP text encoder already provides a meaningful and structured semantic space for LULC concepts. Building upon this fixed space, the renaming trick broadens the semantic coverage of category prompts and improves cross-taxonomy prompt consistency. At the same time, it preserves the distinctions among subtly different class descriptions. This behavior supports our claim that the renaming trick facilitates semantic harmonization at the prompt level.}

\begin{table*}[htp]
\small
\centering
\caption{\cy{Selected urban-related original class names and their renamed prompts used for the text-embedding visualization.}}
\setlength{\extrarowheight}{0.3mm}
\setlength\tabcolsep{3.2pt}
\begin{tabular}{p{2.4cm}p{4cm}p{9.5cm}}
\toprule
\textbf{Dataset} & \textbf{Original name} & \textbf{Renamed prompt} \\
\midrule
OpenEarthMap & Developed space & developed area except for building; built-up area except for building \\
\hline
OpenEarthMap & Road & road; transportation \\
\hline
OpenEarthMap & Building & building \\
\hline
DynamicEarthNet & impervious surface & developed impervious area; built-up impervious area \\
\hline
Iran & Urban & urban area; developed area; residential, commercial, industrial and transportation area; built-up area \\
\hline
GHSL & built spaces, residential & residential area \\
\hline
GHSL & built spaces, non-residential & non-residential built-up area; non-residential developed area; commercial, industrial and transportation area \\
\hline
WorldCover & Built-up & built-up area without artificial vegetation; man-made structure without artificial vegetation; residential, commercial, industrial and transportation area without artificial vegetation; developed area without artificial vegetation \\
\hline
NLCD-LC & Developed, low intensity & developed low-intensity impervious area; built-up low-intensity impervious area \\
\hline
NLCD-LC & Developed, high intensity & developed high-intensity impervious area; built-up high-intensity impervious area \\
\hline
USFS-LU & Developed & developed area; urban area; residential, commercial, industrial and transportation area including artificial vegetation; built-up area \\
\bottomrule
\end{tabular}
\label{app:tab:data:urban_name}
\end{table*}

\begin{table*}[htp]
\small
\centering
\caption{\cy{Selected crop-related original class names and their renamed prompts used for the text-embedding visualization.}}
\setlength{\extrarowheight}{0.3mm}
\setlength\tabcolsep{3.2pt}
\begin{tabular}{p{2.4cm}p{4cm}p{9.5cm}}
\toprule
\textbf{Dataset} & \textbf{Original name} & \textbf{Renamed prompt} \\
\midrule
OpenEarthMap & Agriculture land & agricultural land; crop; crop land \\
DynamicEarthNet & agriculture & agricultural land; crop; crop land \\
Iran & Farm Land & agricultural land; crop; cropland; arable land and permanent crop; herbaceous crop and woody crop; annual crop and orchard and vineyard \\
WorldCover & Cropland & agricultural land except for woody crop; crop except for woody crop; cropland except for woody crop; arable land; herbaceous crop; annual crop \\
NLCD-LC & Cultivated crops & agricultural land; crop; cropland; arable land and permanent crop; herbaceous crop and woody crop; annual crop and orchard and vineyard \\
USFS-LC & Agriculture & agricultural land; crop; cropland; arable land and permanent crop; herbaceous crop and woody crop; annual crop, orchard and vineyard \\
SBTN & crop & agricultural land; crop; cropland; arable land and permanent crop; herbaceous crop and woody crop; annual crop, orchard and vineyard \\
\bottomrule
\end{tabular}
\label{app:tab:data:crop_name}
\end{table*}

\begin{table*}[htp]
\small
\centering
\caption{\cy{Selected vegetation-related original class names and their renamed prompts used for the text-embedding visualization.}}
\setlength{\extrarowheight}{0.3mm}
\setlength\tabcolsep{3.2pt}
\begin{tabular}{p{2.4cm}p{4cm}p{9.5cm}}
\toprule
\textbf{Dataset} & \textbf{Original name} & \textbf{Renamed prompt} \\
\midrule
OpenEarthMap & Rangeland & rangeland and herbaceous wetland; grass, pasture, scrub and herbaceous wetland; herbaceous vegetation, shrub and herbaceous wetland; grass, pasture and shrub and herbaceous wetland; herbaceous vegetation, scrub and herbaceous wetland \\
OpenEarthMap & Tree & tree and woody wetland \\
DynamicEarthNet & forest and other vegetation & forest, grass, shrub, pasture and artificial vegetation; tree, herbaceous vegetation and scrub and artificial vegetation; tree, herb, shrub and artificial vegetation; tree, grass, shrub, pasture and artificial vegetation; forest, herbaceous vegetation, scrub and artificial vegetation; forest, herb, shrub and artificial vegetation \\
DynamicEarthNet & wetland & wetland \\
Iran & Forest & tree; forest; wood; broadleaf and coniferous forest; deciduous and evergreen forest; broadleaf and coniferous tree; deciduous and evergreen tree \\
WorldCover  & Tree cover & tree including corresponding artificial vegetation and woody crop but except for mangroves; forest including corresponding artificial vegetation and woody crop but except for mangroves; wood including corresponding artificial vegetation and woody crop but except for mangroves \\
WorldCover  & Shrubland & shrub and scrub including corresponding artificial vegetation and woody crop; shrub including corresponding artificial vegetation and woody crop; scrub including corresponding artificial vegetation and woody crop \\
\bottomrule
\end{tabular}
\label{app:tab:data:vege_name}
\end{table*}

\clearpage

\subsection{Statistics}

Below, we give the central wavelength information serving as the input of DOFA, along with the band-wise mean and standard deviation (std) values used for data normalization for the LAS dataset.

\begin{table*}[htp]
\centering
\footnotesize
\caption{Data types and band statistics of LAS subsets. \textsuperscript{†}HR: high-resolution, OEM: OpenEarthMap, DEN: DynamicEarthNet, WC: WorldCover.}
\setlength{\extrarowheight}{0.3mm}
\setlength\tabcolsep{6pt}
%
\label{app:tab:data:wave}
\end{table*}

\begin{table*}[htp]
\centering
\footnotesize
\caption{Band-wise mean and standard deviation (std) values of the images in the LAS subsets. The same abbreviations are applied here as in \cref{app:tab:data:wave}.}
\setlength{\extrarowheight}{0.3mm}
\setlength\tabcolsep{6pt}
\begin{tabular}{p{.5cm}<{\centering}|p{.8cm}<{\centering}|p{6.5cm}|p{6.5cm}}
\toprule
\multicolumn{2}{c|}{\textbf{subset}}                                     & \textbf{mean}                                                                                                                                                                                                                                                                                                                                                                                                        & \textbf{std}\\
\midrule
\multirow{2}{*}{HR} & OEM                & {[}123.68, 116.28, 103.53{]}                                                                                                                                                                                                                                                                                                                                                                                & {[}58.395, 57.12, 57.375{]}                                                                                                                                                                                                                                                                                 \\
\bottomrule
\end{tabular}
\label{app:tab:data:meanstd}
\end{table*}

\clearpage
\section{Test sets}
\subsection{Class systems and text prompts}
Below we list the class systems and name strings used for test sets in our experiments.

\begin{table*}[htp]
\small
\caption{Class information of the Potsdam test set. The \textcolor{blue}{first name text prompt} is used in fine-tuning experiments.}
\begin{tabular}{c|c|c|p{10.2cm}}
\toprule
\textbf{Class} & \textbf{Color}    & \textbf{Original name}      & \textbf{Name text prompts for zero-shot inference}                                                                                          \\
\midrule
0     & \cellcolor[HTML]{440154}\textcolor{white}{\#440154} & background         &  \textcolor{blue}{barren land}; bare land; water; crop; agricultural land; rock, sand, clay and soil; wetland; ice and snow \\
\hline
1     & \cellcolor[HTML]{99FFCC}\#99FFCC & low vegetation     & \textcolor{blue}{grass}, shrub and scrub; herb; scrub; shrub; grass                                                   \\
\hline
2     & \cellcolor[HTML]{808080}\#808080 & impervious surface & \textcolor{blue}{road}; transportation; impervious area except for building                                                  \\
\hline
3     & \cellcolor[HTML]{E86ACD}\#E86ACD & car                & \textcolor{blue}{car}                                                                                                        \\
\hline
4     & \cellcolor[HTML]{FFFF19}\#FFFF19 & building           & \textcolor{blue}{building}                                                                                                   \\
\hline
5     & \cellcolor[HTML]{008000}\#008000 & tree               & \textcolor{blue}{tree}; forest     \\                 
\bottomrule
\end{tabular}
\label{app:tab:tsdata:potsdam}
\end{table*}

\begin{table*}[htp]
\small
\caption{Class information of the LoveDA~\citep{wang_loveda_2022} test set. The \textcolor{blue}{first name text prompt} is used in fine-tuning experiments.}
\begin{tabular}{c|c|c|p{11cm}}
\toprule
\textbf{Class} & \textbf{Color}    & \textbf{Original name}      & \textbf{Name text prompts for zero-shot inference}         \\
\midrule
0     & \#FFFFFF & background    & \textcolor{blue}{impervious area except for road and building}; grass; shrub; scrub \\
\hline
1     & \cellcolor[HTML]{FF0000}\textcolor{white}{\#FF0000}   & building      & \textcolor{blue}{building}                                                          \\
\hline
2     & \cellcolor[HTML]{FFFF19}\#FFFF19 & road          & \textcolor{blue}{road}; transportation                                              \\
\hline
3     & \cellcolor[HTML]{0000FF}\textcolor{white}{\#0000FF} & water         & \textcolor{blue}{water}                                                             \\
\hline
4     & \cellcolor[HTML]{7030A0}\textcolor{white}{\#7030A0} & barren        & \textcolor{blue}{barren land}; bare land; soil; rock, sand, clay and soil           \\
\hline
5     & \cellcolor[HTML]{00FF00}\#00FF00 & forest        & \textcolor{blue}{tree}; forest                                                      \\
\hline
6     & \cellcolor[HTML]{E97132}\#E97132 & agriculture   & \textcolor{blue}{agricultural land}; crop; crop land           \\
\bottomrule
\end{tabular}
\label{app:tab:tsdata:loveda}
\end{table*}

\begin{table*}[htp]
\small
\caption{Class information of the NYC (New York City)~\citep{albrecht_monitoring_2022} test set. The \textcolor{blue}{first name text prompt} is used in fine-tuning experiments.}
\begin{tabular}{c|c|p{3cm}<{\centering}|p{9.5cm}}
\toprule
\textbf{Class} & \textbf{Color}    & \textbf{Original name}    & \textbf{Name text prompts for zero-shot inference}                  \\
\midrule
0     & \cellcolor[HTML]{00823C} \#00823C & tree canopy      & \textcolor{blue}{tree}                                                       \\
\hline
1     & \cellcolor[HTML]{9BBB59} \#9BBB59 & grass/shrubs     & \textcolor{blue}{grass}; herb                                                \\
\hline
2     & \cellcolor[HTML]{FF9900} \#FF9900 & bare soil        & \textcolor{blue}{bare land}; barren land; rock, sand, clay and soil          \\
\hline
3     & \cellcolor[HTML]{0000FF}\textcolor{white}{\#0000FF} & water            & \textcolor{blue}{water}                                                      \\
\hline
4     & \cellcolor[HTML]{FFFF19} \#FFFF19 & buildings        & \textcolor{blue}{building}                                                   \\
\hline
5     & \cellcolor[HTML]{5F5F5F} \#5F5F5F & road             & \textcolor{blue}{road except for railway}; transportation except for railway \\
\hline
6     & \cellcolor[HTML]{DDDDDD} \#DDDDDD & other impervious & \textcolor{blue}{impervious area except for road and building}               \\
\hline
7     & \cellcolor[HTML]{C00000} \#C00000 & railroads        & \textcolor{blue}{railway}  \\
\bottomrule
\end{tabular}
\label{app:tab:tsdata:nyc}
\end{table*}

\begin{table*}[htp]
\small
\caption{Class information of the DW (Dynamic world)~\citep{liu_cromss_2025} test set. The \textcolor{blue}{first name text prompt} is used in fine-tuning experiments.}
\begin{tabular}{c|c|p{3cm}<{\centering}|p{9.5cm}}
\toprule
\textbf{Class} & \textbf{Color}    & \textbf{Original name}    & \textbf{Name text prompts for zero-shot inference}             \\
\midrule
0     & \cellcolor[HTML]{0000FF}\textcolor{white}{\#0000FF} & water              & \textcolor{blue}{water}                                                \\
\hline
1     & \cellcolor[HTML]{00823C}\#00823C & trees              & \textcolor{blue}{tree}; forest; wood                                   \\
\hline
2     & \cellcolor[HTML]{9BBB59}\#9BBB59 & grass              & \textcolor{blue}{grass}; herb                                          \\
\hline
3     & \cellcolor[HTML]{99CCFF}\#99CCFF & flooded vegetation & \textcolor{blue}{wetland}                                              \\
\hline
4     & \cellcolor[HTML]{FF9900}\#FF9900 & crops              & \textcolor{blue}{crop}; agricultural land                              \\
\hline
5     & \cellcolor[HTML]{440154}\textcolor{white}{\#440154} & shrub \& scrub     & \textcolor{blue}{shrub and scrub}; shrub; scrub                        \\
\hline
6     & \cellcolor[HTML]{C00000}\#C00000 & built area         & \textcolor{blue}{built-up area}; urban area; developed impervious area \\
\hline
7     & \cellcolor[HTML]{FFFF19}\#FFFF19 & bareland           & \textcolor{blue}{bare land}; barren land; rock, sand, clay and soil    \\

\hline
8     & \cellcolor[HTML]{DDDDDD}\#DDDDDD & ice and snow       & \textcolor{blue}{ice and snow}        \\
\bottomrule
\end{tabular}
\label{app:tab:tsdata:dw}
\end{table*}

\begin{table*}[htp]
\small
\caption{Class information of the OSM (OpenStreetMap)~\citep{liu_cromss_2025} test set. The \textcolor{blue}{first name text prompt} is used in fine-tuning experiments.}
\begin{tabular}{c|c|p{6cm}<{\centering}|p{6.5cm}}
\toprule
\textbf{Class} & \textbf{Color}                   & \textbf{Original name}                   & \textbf{Name text prompts for zero-shot inference} \\
\midrule
0              & \cellcolor[HTML]{C00000}\textcolor{white}{\#C00000} & Urban fabric                             & \textcolor{blue}{residential area}                                   \\
\hline
1              & \cellcolor[HTML]{FF9900}\#FF9900 & Arable land                              & \textcolor{blue}{arable land}                                        \\
\hline
2              & \cellcolor[HTML]{00823C}\#00823C & Forest                                   & \textcolor{blue}{forest}; tree                                       \\
\hline
3              & \cellcolor[HTML]{FF00FF}\#FF00FF & Industrial, commercial \& transport      & \textcolor{blue}{industrial, commercial and transportation area}     \\
\hline
4              & \cellcolor[HTML]{CCFFCC}\#CCFFCC & Artificial, non-agricultural vegetation & \textcolor{blue}{artificial, non-argricultural vegetation}           \\
\hline
5              & \cellcolor[HTML]{DDDDDD}\#DDDDDD & Mine, dump \&  construction              & \textcolor{blue}{mine, dump and construction site}                   \\
\hline
6              & \cellcolor[HTML]{9BBB59}\#9BBB59 & Pastures                                 & \textcolor{blue}{pasture}                                            \\
\bottomrule
\end{tabular}
\label{app:tab:tsdata:osm}
\end{table*}

\begin{table*}[htp]
\small
\caption{Class information of the MultiSenGe~\citep{wenger_multimodal_2023} test set. The \textcolor{blue}{first name text prompt} is used in fine-tuning experiments.}
\begin{tabular}{c|c|p{6cm}<{\centering}|p{6.5cm}}
\toprule
\textbf{Class} & \textbf{Color}                   & \textbf{Original name}           & \textbf{Name text prompts for zero-shot inference} \\
\midrule
0              & \cellcolor[HTML]{FF5149}\#FF5149 & Dense Built-Up                   & \textcolor{blue}{built-up high-intensity impervious area}            \\
\hline
1              & \cellcolor[HTML]{FF9900}\#FF9900 & Sparse Built-Up                  & \textcolor{blue}{built-up low-intensity impervious area}             \\
\hline
2              & \cellcolor[HTML]{D86DCD}\#D86DCD & Specialized Built-Up Areas       & \textcolor{blue}{industrial and commercial area}                     \\
\hline
3              & \cellcolor[HTML]{9BBB59}\#9BBB59 & Specialized but Vegetative Areas & \textcolor{blue}{artificial, non-argricultural vegetation}           \\
\hline
4              & \cellcolor[HTML]{DDDDDD}\#DDDDDD & Large Scale Networks             & \textcolor{blue}{transportation; road}                               \\
\hline
5              & \cellcolor[HTML]{FFFF99}\#FFFF99 & Arable Land                      & \textcolor{blue}{arable land; herbaceous crop}                       \\
\hline
6              & \cellcolor[HTML]{FFCCFF}\#FFCCFF & Vineyards                        & \textcolor{blue}{vineyard}                                           \\
\bottomrule
\end{tabular}
\label{app:tab:tsdata:multisenge}
\end{table*}

\clearpage
\section{\cy{Regional-Scale Visualization Results}}

\begin{figure*}[htp]
    \centering
    \includegraphics[width=.9\linewidth]{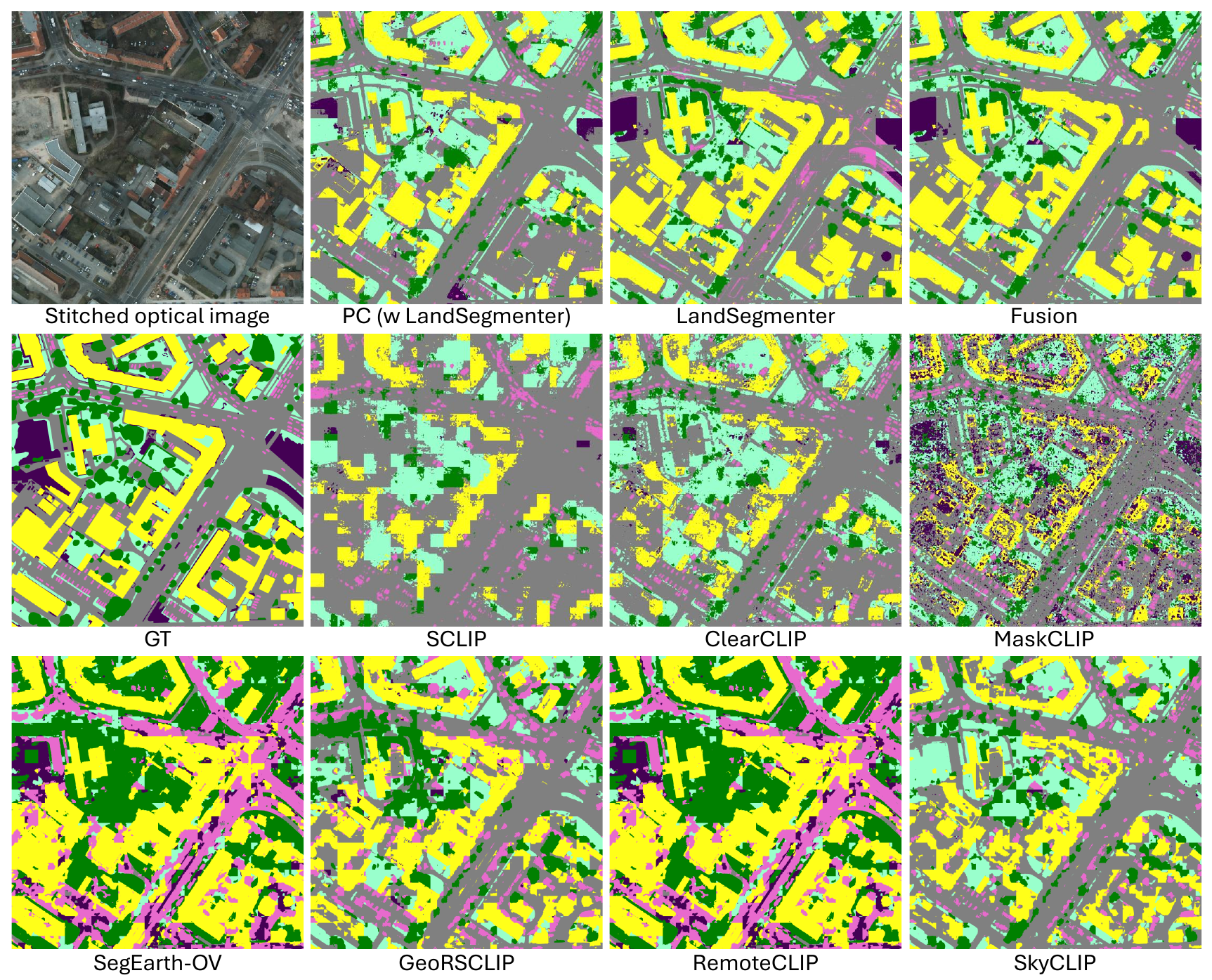}
    \caption{\cy{Regional-scale qualitative comparison on the Potsdam dataset. The scene is constructed by stitching predictions from an $11 \times 11$ patch region.}}
    \label{app:fig:region_potsdam}
\end{figure*}

\begin{figure*}[htp]
    \centering
    \includegraphics[width=.9\linewidth]{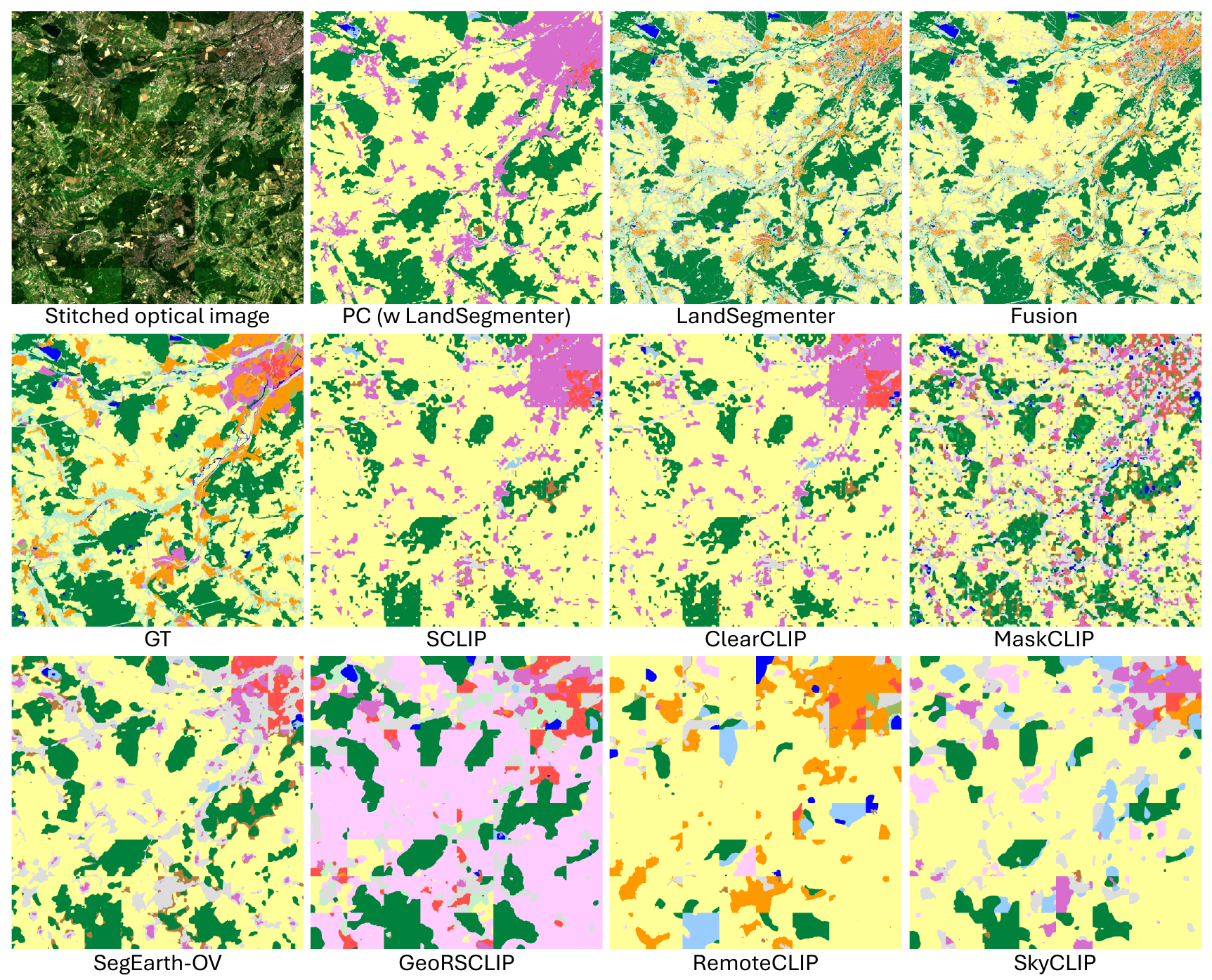}
    \caption{\cy{Regional-scale qualitative comparison on the MultiSenGE dataset. The scene is constructed by stitching predictions from an $8 \times 8$ patch region of Sentinel-2 imagery.}}
    \label{app:fig:region_multisenge}
\end{figure*}

\cy{To further evaluate the spatial consistency of the proposed method beyond the cropped examples shown in the main manuscript, we provide two larger-scale qualitative visualization results obtained under the zero-shot setting in this appendix. The first scene is from the Potsdam dataset, which contains very high-resolution imagery at 0.05m. As shown in \cref{app:fig:region_potsdam}, we stitched predictions from an $11 \times 11$ patch region to obtain a broader urban scene for this case. The second scene is from the MultiSenGe dataset, which is based on Sentinel-2 imagery at 10m resolution. As illustrated in \cref{app:fig:region_multisenge}, we stitched predictions from an $8 \times 8$ patch region to visualize a larger regional context. These two examples were selected because they represent two different resolution regimes and provide reliable spatial coordinates for scene stitching.}

\cy{The regional-scale zero-shot results further highlight the advantages of the proposed methods over other comparison approaches. In the high-resolution Potsdam scene, our method shows a more complete semantic understanding of the urban layout, yielding more coherent predictions for roads, buildings, and vegetation while preserving finer structural details and object boundaries. In the lower-resolution MultiSenGe scene, our method also produces more spatially consistent large-area land-cover patterns, with less severe semantic confusion and better spatial details than the other methods. Overall, these visualizations suggest that the proposed framework not only improves global semantic coherence, but also retains stronger detail characterization ability across different spatial resolutions.}

\end{document}